\documentclass[dvipsnames]{article}

\usepackage[final]{lychee}
\usepackage[utf8]{inputenc} 
\usepackage[T1]{fontenc}    
\usepackage{url}            
\usepackage{booktabs}       
\usepackage{amsfonts}       
\usepackage{nicefrac}       
\usepackage{booktabs} 
\usepackage{multirow}
\usepackage{wrapfig}
\usepackage{amssymb}
\usepackage{epigraph}
\usepackage{makecell}
\usepackage{listings}
\usepackage{fvextra}
\usepackage{subcaption}
\usepackage{algorithm}
\usepackage{algorithmic}
\usepackage[labelfont=bf]{caption} 

\usepackage{color, soul}
\usepackage{microtype}      
\usepackage{xcolor}         
\usepackage{geometry}
\usepackage{mathpazo}
\usepackage[xcharter,bigdelims,vvarbb,noamssymbols,nosymbolsc]{newtxmath}
\usepackage[scaled=1.1]{zlmtt}
\usepackage{ragged2e}
\usepackage{amsmath}
\usepackage{bm}
\usepackage{latexsym}
\usepackage{graphicx}
\usepackage{float}
\usepackage{longtable}
\usepackage{booktabs} 
\usepackage{multirow}
\usepackage{amssymb}
\usepackage{longtable}
\usepackage{tabu}
\usepackage{bbding}
\usepackage{bbding}
\usepackage[most,breakable]{tcolorbox}
\usepackage{etoolbox}
\usepackage{fancyhdr}
\usepackage{lipsum}  
\usepackage{CJKutf8}
\usepackage{pdfpages}
\usepackage{multicol}
\usepackage{pgffor} 
\usepackage{nicematrix} 
\usepackage{paralist}
\usepackage[edges]{forest}
\usepackage{siunitx}
\usepackage{tikz-dependency}
\usepackage{tikz}
\usepackage{bbm}
\usepackage{amssymb}
\usepackage[english]{babel}  
\usepackage{hyphenat}
\usepackage{siunitx}

\newcommand{\Rmnum}[1]{\expandafter\@slowromancap\romannumeral #1@}

\usepackage{caption} 
\usepackage{chngcntr}

\usepackage[colorlinks, linkcolor=RoyalBlue, anchorcolor=BrickRed, citecolor=RoyalBlue, urlcolor=RoyalBlue]{hyperref}

\usepackage{cleveref}
\crefname{section}{§}{§§}
\Crefname{section}{§}{§§}

\newcommand\refsec[1]{Section~\hyperref[sec:#1]{\ref{sec:#1}}}
\newcommand\refsecs[2]{\hyperref[sec:#1]{§\ref{sec:#1}:~\textsc{#1}}, \hyperref[sec:#2]{§\ref{sec:#2}:~\textsc{#2}}}

\definecolor{msftBlue}{RGB}{0,164,239}
\definecolor{msftGreen}{RGB}{127,186,0}
\definecolor{msftYello}{RGB}{255,185,0}
\definecolor{mypurple}{RGB}{138,43,226} 
\definecolor{msftBlack}{RGB}{0,0,0}

\newtcolorbox{myboxnote}[1][]{
  breakable,
  title=#1,
  colback=cyan!0,
  colbacktitle=cyan!0,
  coltitle=black,
  fonttitle=\bfseries,
  bottomrule=0pt,
  toprule=0pt,
  leftrule=1.5pt,
  rightrule=1.5pt,
  titlerule=0pt,
  arc=0pt,
  outer arc=0pt,
  colframe=lightgray,
}
\usepackage{mdframed}

\definecolor{academicblue}{RGB}{54, 95, 145}

\tcbset{
  iclrtakeawaybox/.style={
    width=\linewidth,
    colback=gray!5,                 
    colframe=gray!60,              
    colbacktitle=gray!30,            
    coltitle=black,                
    fonttitle=\bfseries,     
    boxrule=0.5pt,                 
    arc=2pt,                       
    sharp corners=south,           
    attach boxed title to top left={yshift=-0.1in,xshift=0.15in},
    boxed title style={boxrule=0pt, colframe=white},
    enhanced,
    drop shadow southeast           
  }
}
\newtcolorbox{TakeawayBox}[2][]{iclrtakeawaybox,title=#2,#1}

\definecolor{promptframegray}{RGB}{124,124,124}
\definecolor{promptbodygray}{RGB}{249,249,249}
\newtcolorbox{PromptBox}[2][]{
  enhanced jigsaw,
  breakable,
  width=\linewidth,
  colback=promptbodygray,
  colframe=promptframegray,
  colbacktitle=promptframegray,
  coltitle=white,
  fonttitle=\fontfamily{phv}\selectfont\bfseries,
  fontupper=\small,
  title={#2},
  boxrule=0.7pt,
  arc=3pt,
  outer arc=3pt,
  boxsep=0pt,
  left=7pt,
  right=7pt,
  top=5pt,
  bottom=5pt,
  lefttitle=7pt,
  righttitle=7pt,
  toptitle=4pt,
  bottomtitle=4pt,
  before skip=6pt,
  after skip=8pt,
  before upper={%
    \setlength{\parindent}{0pt}%
    \setlength{\parskip}{3pt}%
  },
  #1
}

\providecommand{\tightlist}{%
  \setlength{\itemsep}{4pt}%
  \setlength{\parskip}{0pt}%
}
\newcommand{\PromptHeading}[1]{%
  \par\smallskip
  \noindent{\bfseries #1}\par
  \smallskip
}

\newenvironment{itemsize*}%
 {\leftmargini=20pt\begin{itemize}%
  \setlength{\itemsep}{3pt}%
  \setlength{\parskip}{0pt}%
  }%
 {\end{itemize}}
\newenvironment{enumerate*}%
 {\begin{enumerate}%
  \setlength{\itemsep}{0pt}%
  \setlength{\parskip}{0pt}}%
 {\end{enumerate}}

\usepackage{fancyhdr} 
\usepackage{blindtext} 
\usepackage{ulem}

\usepackage{xparse}
\usepackage{colortbl}

\usepackage{svg}

\newcommand{\methodfull}{\textsc{LycheeMemory V2}}
\newcommand{\method}{\textsc{LycheeMemory}}

\title{
\vspace{-2em}
\fontsize{16}{19}\selectfont \methodfull{}: Efficient Long-Term Memory for LLM Agents via Semantic Segment-Level Consolidation}
\author{
Dongfang Li\textsuperscript{1}\thanks{~Equal contribution.},
Zixuan Liu\textsuperscript{1}\footnotemark[1],
Junmai Wang\textsuperscript{1},
Jiahe Huang\textsuperscript{1},\\
Fuhao Li\textsuperscript{1},
Lin Zhang\textsuperscript{2},
Bonian Jia\textsuperscript{1},
Baotian Hu\textsuperscript{1}\thanks{Corresponding Author},
Min Zhang\textsuperscript{1}\\
\textsuperscript{1}Harbin Institute of Technology, Shenzhen\\
\textsuperscript{2}Huawei\\
}

\begin{document}

\maketitle
\vspace{-1em}

\begin{abstract}
Long-horizon LLM agents must preserve information from past interactions to support future tasks. Existing memory systems typically rely on eager consolidation, invoking LLMs after each interaction to extract, summarize, or update memories. This design makes memory construction increasingly costly as conversations grow. Coarse summarization can reduce construction cost but risks discarding fine-grained contextual evidence, whereas larger retrieval contexts or multi-hop LLM reasoning shift the overhead to query time. We present \methodfull{}, an efficient long-term memory framework that replaces turn-level consolidation with semantic segment-level consolidation. Instead of consolidating every interaction, \method{} batches multiple exchanges into segments and encodes each finalized segment into context-independent typed memory records. Segment-level batching lowers LLM encoding frequency, while semantic boundary detection helps preserve coherent event-level and temporal evidence compared with fixed-window batching. The resulting records are organized with lightweight structured indexes for query-planned evidence retrieval. Experiments using \texttt{GPT-4.1-Mini} show that \method{} achieves state-of-the-art performance, reaching 89.22\% on LoCoMo and 92.20\% on LongMemEval-S. Compared with A-Mem, it reduces construction tokens by 86.0\% on LoCoMo and 75.9\% on LongMemEval-S without increasing query-time token usage. More broadly, our results suggest that the accuracy--cost trade-off of long-term agent memory depends not only on what information is retained, but also on the granularity at which it is consolidated.
\end{abstract}

\begin{center}
    \includegraphics[width=\linewidth]{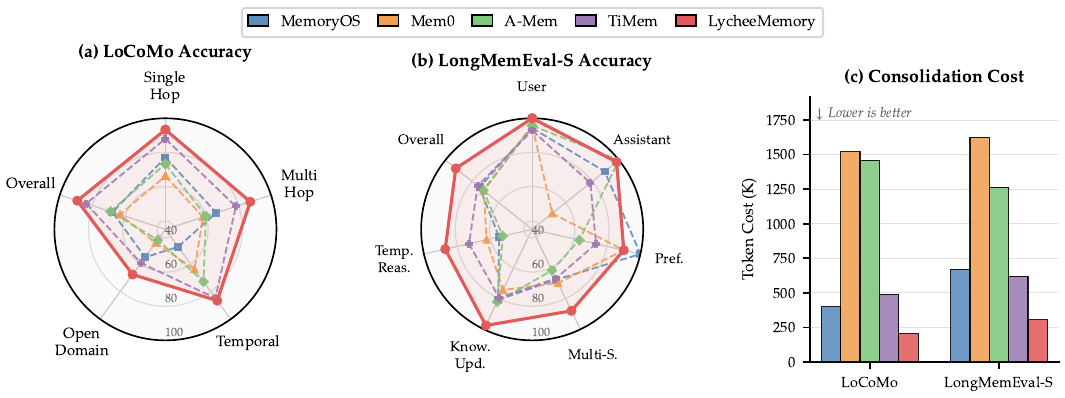}
    \captionof{figure}{\method{} achieves superior accuracy with substantially lower consolidation cost under \texttt{GPT-4.1-Mini}. \textbf{Left:} Category-wise accuracy comparison on LoCoMo, covering single-hop, multi-hop, temporal, open-domain, and overall questions. \textbf{Middle:} Category-wise accuracy comparison on LongMemEval-S, including user, assistant, preference, multi-session, knowledge-update, temporal-reasoning, and overall questions. \textbf{Right:} Memory consolidation cost comparison across LoCoMo and LongMemEval-S, where lower is better.}
    \label{fig:main_results}
\end{center}

\newpage
{
  \hypersetup{linkcolor=RoyalBlue, linktoc=page}
  \tableofcontents
}

\newpage

\section{Introduction}
\label{sec:introduction}

Large language model (LLM) agents \citep{memorybank, memgpt, memobase} are increasingly expected to operate as persistent assistants rather than single-turn responders. In long-horizon settings \citep{locomo, longmemeval} such as personal assistance, customer support, tutoring, and task-oriented dialogue, an agent \citep{memorybank, mem0} must remember user preferences, historical events, evolving facts, constraints, and previous failures across many interactions.

Since an LLM cannot reliably keep an unbounded interaction history in its active context window, recent systems \citep{memgpt, memoryos, mem0, a-mem, g-long, horma} have introduced external memory modules to extract, update, organize, and retrieve information from past conversations. A common design in these systems \citep{mem0, a-mem, memoryos, memrouter, memrefine} is eager consolidation: after each turn or short exchange, an LLM is invoked to summarize the interaction, extract facts, update existing memories, or build links with historical memories. Although effective, this turns memory construction into a high-frequency LLM operation \citep{memrouter, g-long, memrefine, dmf} whose token cost accumulates rapidly as conversations grow.

Simply making memory cheaper by storing coarser summaries is also insufficient, because long-term memory questions \citep{locomo, longmemeval} often depend on fine-grained evidence \citep{g-long, horma, evimem} such as entities, temporal expressions, coreference relations, and small contextual details. Meanwhile, improving recall \citep{ember} by expanding top-$k$ retrieval or using LLM-driven multi-hop search \citep{evimem, memflow, memreranker} introduces additional query-time overhead. Thus, the key challenge is not merely how to store more history, but how to preserve sufficient evidence at the right granularity while keeping both construction and retrieval costs under control.

\begin{figure*}[t]
\centering
\includegraphics[width=\textwidth]{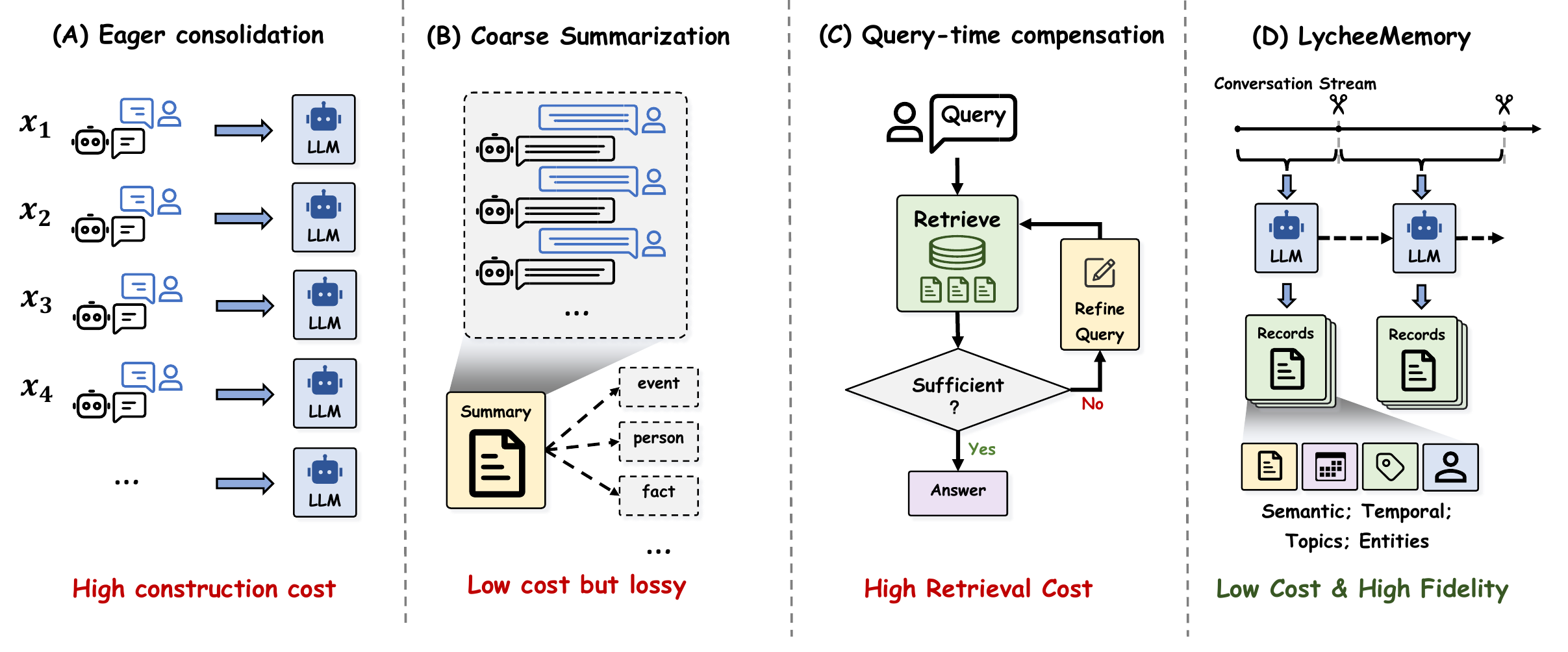}
\caption{Motivation of \method{}. \textbf{(A)} Eager turn-level consolidation incurs high construction cost. \textbf{(B)} Coarse summarization reduces construction cost but may discard fine-grained evidence. \textbf{(C)} Iterative query-time compensation shifts overhead to retrieval. \textbf{(D)} \method{} batches multiple exchanges into semantically coherent segments and encodes each finalized segment into typed evidence records for structured retrieval, reducing construction frequency while preserving fine-grained evidence without query-time expansion.}
\label{fig:motivation}
\end{figure*}

Motivated by these limitations, we propose \methodfull{}, a cost-efficient long-term memory framework for LLM agents (Figure~\ref{fig:motivation}). Instead of consolidating every turn, \method{} batches multiple exchanges and invokes the LLM once for each finalized segment, thereby reducing construction frequency. Within this segment-level batching scheme, semantic surprise and cohesion signals determine where segments end, helping preserve coherent event boundaries rather than relying on mechanical fixed windows. Each segment is then encoded into context-independent typed memory records that contain natural-language statements, memory types, entities, topics, temporal scopes, and links to the original dialogue evidence. To maintain continuity across segments without re-reading the full history, the system carries lightweight disambiguation feedback, including entity aliases and reference relations, into later consolidation steps. \method{} further organizes records in an append-only structured memory store with entity, topic, temporal, event-frame, and entity-topic indexes. At query time, a planner decomposes the user question into multiple recall routes and retrieves evidence from semantic memory, structured indexes, temporal indexes, and raw turns, followed by route-level reranking, fusion, and diversity-aware selection.

We evaluate \method{} on LoCoMo and LongMemEval-S \citep{locomo, longmemeval}. The results show that \method{} consistently improves long-term memory QA accuracy while substantially reducing construction-token cost, achieving over 20-point accuracy gains over A-Mem \citep{a-mem} with up to 7.2$\times$ fewer construction tokens. These improvements are obtained with lower query-time token usage than A-Mem, with particularly large gains on multi-hop, temporal-reasoning, preference-tracking, and multi-session questions.

Our contributions are threefold:
\begin{itemize}

\item We identify memory-construction granularity as an important efficiency lever for long-term agent memory: batching multiple exchanges reduces write-side LLM calls, while semantic boundaries preserve coherent evidence relative to fixed windows.

\item We introduce \method{}, a semantic segment-level memory construction framework that encodes each finalized segment into typed, self-contained records. Bounded cross-segment context maintains entity and reference continuity despite less frequent consolidation.

\item We demonstrate a stronger accuracy--cost trade-off on LoCoMo and LongMemEval-S, improving long-term memory QA accuracy while reducing construction-token cost and query-time token usage relative to A-Mem. Ablations further distinguish the cost effect of segment-level batching from the accuracy effect of semantic boundary selection.

\end{itemize}

\section{Related Work}
\label{sec:related_work}

\subsection{Long-Term Memory for LLM Agents}

Long-term memory has become a central component of LLM agents that must operate across sessions rather than within a single context window. Early systems \citep{memorybank, memgpt, memobase} introduced external memory to preserve user facts, preferences, and historical context, thereby moving from stateless response generation toward persistent assistance. More recent frameworks broaden this agenda by separating different forms of memory or by coupling memory with agent control. For example, MemoryOS \citep{memoryos} and A-Mem \citep{a-mem} emphasize richer memory abstractions and organization, while G-Long \citep{g-long} and HORMA \citep{horma} connect memory with more structured reasoning or navigation mechanisms. In parallel, benchmarks such as LoCoMo \citep{locomo} and LongMemEval \citep{longmemeval} clarify that long-term memory quality depends not only on whether information is retained, but also on whether it can be reconstructed and used under long-horizon budget constraints.

Taken together, this literature suggests that long-term memory systems are increasingly differentiated along three technical axes: how memories are constructed from interaction streams, how they are organized and retrieved for downstream reasoning, and how cost is controlled during long-horizon use. We organize the remainder of this section around these three axes.

\subsection{Memory Construction for LLM Agents}

A dominant line of work constructs memory eagerly from incoming dialogue. Mem0 \citep{mem0} extracts and updates memories at turn granularity, while A-Mem \citep{a-mem} enriches this pattern with linked notes and more structured memory units. MemoryOS \citep{memoryos} and related frameworks \citep{nemori, toki} introduce stronger memory hierarchies or management policies, but still rely on LLM-mediated ingestion or summarization as new dialogue arrives. The common strength of this family is semantic richness at write time; its main drawback is that memory construction remains a frequent LLM operation.

A second line of work reduces write-side cost after or around this basic pipeline. MemRefine \citep{memrefine} treats compression as a post-construction decision problem, using the LLM to delete, merge, or preserve entries under a budget. DMF \citep{dmf} removes LLM calls from memory management through deterministic scoring and decay, trading semantic flexibility for lower and more reproducible cost. MemRouter \citep{memrouter} makes a different trade-off by learning an embedding-based admission policy that decides whether a new interaction should enter memory. These methods show that construction can be made cheaper in different ways, but most of them either optimize memory after frequent writes have already been produced or make storage decisions at relatively fine temporal granularity.

Recent systems have also explored segment-level memory construction. SeCom \citep{secom} partitions conversations into topically coherent segments and compresses them before retrieval, while HiMem \citep{himem} uses topic-aware event-surprise segmentation to construct episode memories within a hierarchical memory architecture.

\method{} shares this segment-level construction granularity but focuses specifically on online write-side efficiency. It uses embedding-based boundary decisions as a construction trigger, allowing multiple exchanges to share one encoding call instead of producing separate turn-level updates. Semantic boundary detection determines the composition of each batch, while bounded cross-segment disambiguation supports the construction of self-contained records.

\subsection{Memory Organization and Retrieval}

Once memory has been written, the next challenge is how to retrieve the right evidence for complex reasoning. Many early systems \citep{memorybank, memobase, mem0} expose memory primarily through flat semantic retrieval over summaries, notes, or profile-like records. This design works well for direct factual lookup, but it is less well aligned with questions that depend on temporal constraints, multi-hop relations, or comparisons across episodes \citep{locomo, longmemeval}, where entity identity, topical scope, and temporal validity must be handled explicitly.

A substantial body of work addresses this problem by adding structure to the memory store itself. G-Long \citep{g-long} and REAL \citep{real} organize memory through structured relations or graph-like representations, while hierarchical or workspace-style systems such as HORMA \citep{horma}, MAGE \citep{mage}, and Infini-Memory \citep{infini-memory} support reasoning across multiple granularities. These methods show that richer structure can improve long-horizon retrieval, especially when evidence must be composed rather than directly matched. At the same time, the amount of extra indexing, maintenance, or traversal machinery varies substantially across systems, so the trade-off is better characterized as additional structural complexity rather than a uniform cost increase.

Another line of work improves retrieval through more explicit query-time control. EviMem \citep{evimem} retrieves evidence iteratively based on missing information, MemFlow \citep{memflow} routes a query to different memory operations, and MemReranker \citep{memreranker} shows that reasoning-aware reranking can substantially improve final evidence quality. These approaches make clear that long-term memory retrieval is often operation-aware rather than single-shot. \method{} is closest in spirit to this structured and routed line, but it emphasizes a lighter query-time design: it uses one planning step to dispatch recall over pre-built indices instead of relying on iterative retrieval loops.

\subsection{Efficient Memory Systems}

Recent memory research increasingly treats efficiency as a primary objective rather than a secondary implementation concern. EMBER \citep{ember} studies budgeted evidence retention, asking which information should remain available under a fixed memory budget. ActiveMem \citep{activemem} reduces the burden on the main model by offloading memory distillation to lighter components, while LightMem \citep{LightMem} develops a lightweight memory-augmented generation pipeline. ScrapMem \citep{scrapmem} focuses on storage-constrained multimodal personalization, while Memanto \citep{memanto} argues that some graph-plus-vector memory designs impose avoidable infrastructure and latency overheads. DMF \citep{dmf} and MemRouter \citep{memrouter} similarly illustrate that large efficiency gains are possible when parts of memory management are replaced by deterministic or lightweight policies.

This literature establishes an important evaluation principle: memory systems should be compared not only by answer quality but also by their token, latency, and storage footprints. \method{} reduces construction-time token cost by lowering the frequency of semantic consolidation.

\section{Method}
\label{sec:method}

\newcounter{methodalg}
\renewcommand{\themethodalg}{\arabic{methodalg}}

\subsection{Overview}
\label{subsec:overview}

We formulate long-term agent memory as an online construction and retrieval problem. Given a multi-session conversation stream $\mathcal{C}=\{x_1,\ldots,x_T\}$, where each exchange $x_t$ contains a user message and the corresponding assistant response, the system incrementally builds a memory store $\mathcal{M}$ and later retrieves evidence for a query $q$. The objective is to maximize answer accuracy while controlling both construction cost, measured by total LLM tokens used during memory building, and query cost, measured by LLM tokens consumed per question.

\method{} replaces turn-level consolidation with semantic segment-level consolidation. Instead of invoking an LLM after every exchange, it buffers multiple exchanges and performs one encoding pass per finalized segment. This segment-level batching reduces construction frequency. A semantic boundary detector separately determines which exchanges belong to each segment so that the batching policy follows conversational structure rather than fixed windows. Each segment is converted into typed, context-independent memory records, enriched with entities, topics, temporal scopes, and provenance links. The records are stored in an append-only evidence store with semantic retrieval and structured indexes for entity, topic, temporal, event-frame, and entity-topic access. At query time, a planner decomposes the question into typed recall routes, and deterministic retrieval modules collect, fuse, and select evidence from the indexed memory store.

\begin{figure*}[t]
\centering
\includegraphics[width=\textwidth]{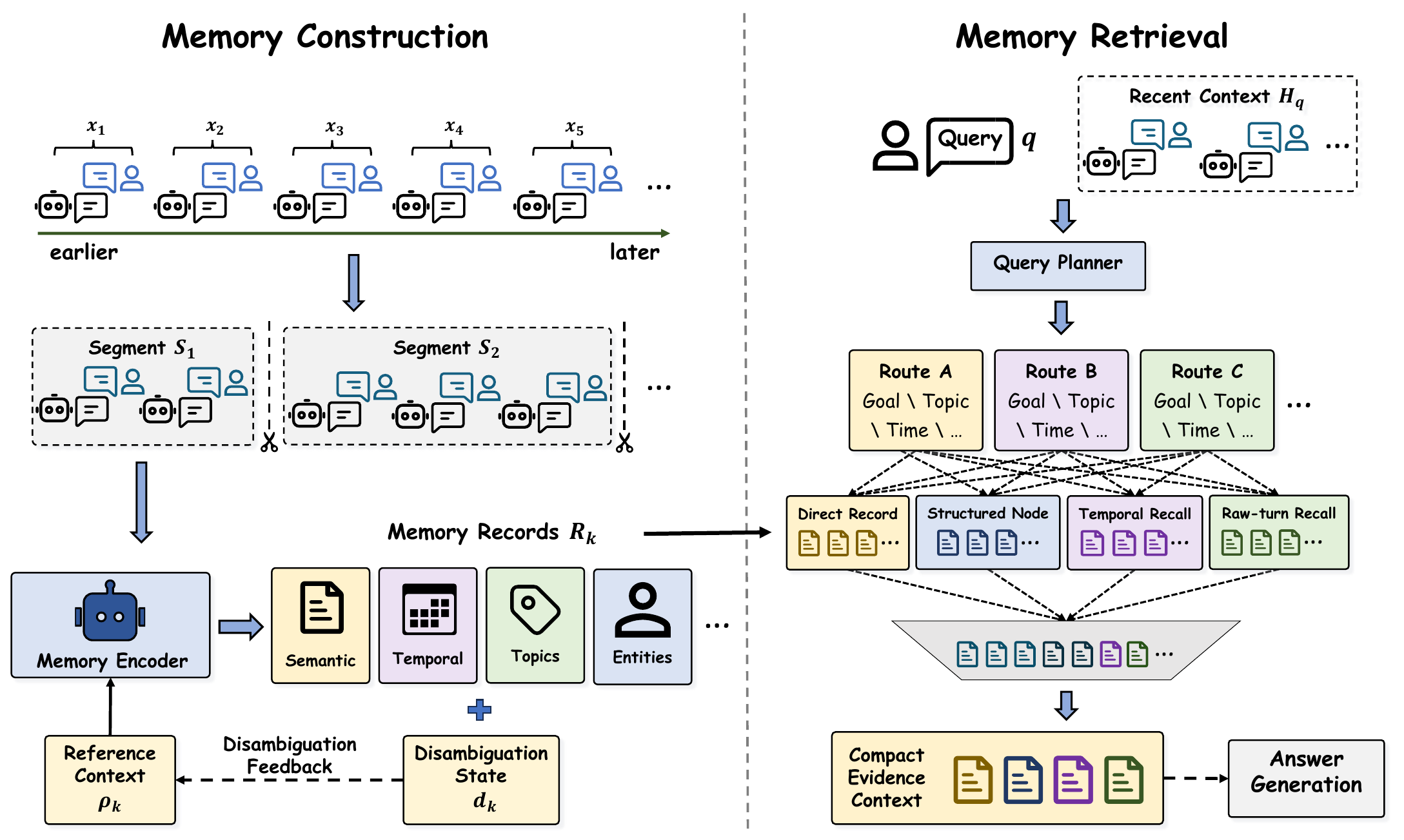}
\caption{Overview of \method{}. During memory construction, online semantic segmentation partitions the conversation stream into coherent segments, and the memory encoder converts each finalized segment into typed records with semantic, temporal, topic, and entity information. A compact reference context and disambiguation feedback maintain continuity across segments. During memory retrieval, the query planner converts the current query and recent context into route-specific evidence needs. Direct-record, structured-node, temporal, and raw-turn recall collect candidates, which are fused into a compact evidence context for answer generation.}
\label{fig:pipeline}
\end{figure*}

Figure~\ref{fig:pipeline} illustrates the construction and retrieval phases. Sections~\ref{subsec:online-semantic-segmentation}--\ref{subsec:structured-evidence-organization} detail memory construction, and Section~\ref{subsec:plan-guided-multi-route-retrieval} describes retrieval. Algorithms~\ref{alg:construction-text} and~\ref{alg:retrieval-text} summarize the two procedures.

\subsection{Online Semantic Segmentation}
\label{subsec:online-semantic-segmentation}

The first component decides when a dialogue fragment is sufficiently complete to be consolidated. Turn-level eager consolidation is costly because it applies the same LLM encoding operation to every exchange, including exchanges that are still part of an unfinished semantic episode. Batching multiple exchanges into one segment reduces this call frequency. Fixed windows provide such batching mechanically, but they ignore event boundaries and may split temporally or referentially coherent evidence. We therefore use online semantic segmentation as the boundary policy within segment-level batching, triggering consolidation when the active segment becomes semantically saturated or a topic transition is detected.

Let $S_k=\{x_s,\ldots,x_{t-1}\}$ denote the current active segment before observing a new exchange $x_t$. We encode each exchange into an embedding vector $\mathbf{e}_t$, maintain the segment centroid $\mathbf{c}_k$, and keep the most recent exchange embedding $\mathbf{h}_k$. The semantic surprise score $s_t$ is defined as
\begin{equation}
s_t
=1-\max\bigl(\mathrm{sim}(\mathbf{e}_t,\mathbf{c}_k),\mathrm{sim}(\mathbf{e}_t,\mathbf{h}_k)\bigr),
\tag{1}
\end{equation}
where $\mathrm{sim}(\cdot,\cdot)$ is cosine similarity. This score captures whether the new exchange departs from both the global topic of the active segment and the local conversational trajectory.

We also measure how much the new exchange weakens internal segment coherence. For a set of exchange embeddings $V$, let
\begin{equation}
\mathrm{Coh}(V)=\frac{1}{|V|}\sum_{\mathbf{v}\in V}\mathrm{sim}(\mathbf{v},\mathbf{c}_V),
\tag{2}
\end{equation}
where $\mathbf{c}_V$ is the centroid of $V$. The cohesion drop $d_t$ induced by $x_t$ is
\begin{equation}
d_t
=\max\bigl(0,\mathrm{Coh}(S_k)-\mathrm{Coh}(S_k\cup\{x_t\})\bigr).
\tag{3}
\end{equation}
A larger drop indicates that adding the exchange would make the segment less topically coherent.

The final boundary score combines semantic surprise, cohesion drop, token pressure, and turn-count pressure:
\begin{equation}
p_t=\sigma\bigl(
b+w_s\phi(s_t)+w_c d_t
+w_l L_t+w_n N_t
\bigr),
\tag{4}
\end{equation}
where $\phi(\cdot)$ applies local normalization to semantic surprise, $L_t$ is the normalized token-length pressure, and $N_t$ is the normalized turn-count pressure. A segment is finalized when $p_t$ exceeds a fixed threshold $\delta$ or a hard token or exchange-count cap is reached; otherwise, $x_t$ is appended to the active segment. This design keeps the boundary decision fully embedding-based, so mid-segment exchanges are buffered without LLM inference. The reduction in encoding calls comes from consolidating at segment rather than exchange granularity: the number of calls is proportional to the number of segments $|\mathcal{S}|$ rather than the number of exchanges $T$. The semantic boundary score determines the composition of those segments.

\subsection{Segment-Level Memory Encoding}
\label{subsec:segment-level-memory-encoding}

The second component converts each finalized segment into memory records that can be retrieved and interpreted without the original dialogue context. Raw turns often contain pronouns, elliptical references, relative dates, and implicit entity mentions. If such turns are stored directly, later retrieval must recover missing context at query time. \method{} instead resolves these dependencies at write time within a coherent segment, where the necessary local context is still available.

For a finalized segment $S_k$, the encoder receives the segment text together with a compact reference context $\rho_k$ from recent segments. It produces a set of memory records $\mathcal{R}_k$ and an updated disambiguation state $d_k$:
\begin{equation}
(\mathcal{R}_k,d_k)=\mathrm{Encode}(S_k,\rho_k).
\tag{5}
\end{equation}
The encoding prompt asks the LLM to perform three operations in a single pass: extract atomic information units, resolve coreference and elliptical mentions, and normalize relative temporal expressions using the session timestamp.

Each record $r_i\in\mathcal{R}_k$ is represented as
\begin{equation}
r_i=\bigl(
\mathrm{id}_i,\tau_i,\mathrm{text}_i,\mathcal{E}_i,\mathcal{K}_i,\mathcal{T}_i,\mathrm{src}_i
\bigr),
\tag{6}
\end{equation}
where $\mathrm{id}_i$ is an internal record identifier, $\tau_i$ is the memory type, $\mathrm{text}_i$ is a self-contained natural-language statement, $\mathcal{E}_i$ is the entity set, $\mathcal{K}_i$ is the topic-tag set, $\mathcal{T}_i$ records normalized event or validity times when available, and $\mathrm{src}_i$ stores provenance links to the source turns. The memory type $\tau_i$ is selected from a finite schema covering facts, preferences, events, constraints, procedures, failure patterns, and tool affordances.

To preserve continuity across segment boundaries, the encoder also returns a disambiguation state $d_k$ containing resolved aliases, canonical entity names, and reference relations that may be needed by later segments. The reference context for the next segment is constructed as
\begin{equation}
\rho_{k+1}=\bigl[d_k;\mathrm{Recent}(\mathcal{R}_{k-m:k})\bigr],
\tag{7}
\end{equation}
where $\mathrm{Recent}(\cdot)$ selects a bounded set of recent record summaries from the same session. This context is truncated to a fixed budget, which prevents the prompt size from growing with the full conversation history. The result is a sequence of records that are locally self-contained while remaining consistent across segments.

\subsection{Structured Evidence Organization}
\label{subsec:structured-evidence-organization}

The third component organizes encoded records so that retrieval can satisfy both semantic and structured constraints. Flat vector search is effective for approximate semantic matching, but long-term memory questions often require explicit access by entity, topic, time, event context, or combinations of these fields. \method{} therefore builds structured evidence indexes directly from the metadata already produced by the segment encoder, without additional LLM calls.

For each record $r_i$, the organizer inserts the record into a vector store and updates a structured store. The vector store embeds $\mathrm{text}_i$ for direct semantic retrieval. The structured store maintains five classes of evidence nodes:
\begin{itemize}
\item entity nodes, one for each entity in $\mathcal{E}_i$;
\item topic nodes, one for each topic tag in $\mathcal{K}_i$;
\item entity-topic nodes, one for each co-occurring pair $(e,k)\in\mathcal{E}_i\times\mathcal{K}_i$;
\item temporal nodes, keyed by day-level or month-level temporal scopes in $\mathcal{T}_i$;
\item event-frame nodes, each grouping the records produced from the same finalized semantic segment while retaining the source session and turn range as provenance.
\end{itemize}
Each evidence node stores a searchable text representation, an optional embedding, and pointers to linked memory records.

The structured store retains textually distinct statements as separate records. Direct record search retrieves records by semantic similarity, while evidence-node search retrieves or filters structured nodes and expands them to linked records. Temporal nodes support exact-date and range-based filtering, and event-frame nodes preserve local co-occurrence among records extracted from the same finalized segment. Evidence distributed across segments or sessions is assembled later by retrieving and fusing records from multiple evidence nodes and recall routes. Because all nodes are derived from record metadata, this phase requires only embedding and bookkeeping operations, and the write-side LLM cost remains dominated by segment-level encoding.

Algorithm~\ref{alg:construction-text} summarizes memory construction. The encoder uses cross-segment reference context to resolve aliases and references while extracting facts from the current segment.

\begin{table*}[t]
\centering
\refstepcounter{methodalg}\label{alg:construction-text}
{\small
\setlength{\tabcolsep}{3pt}
\begin{tabular}{@{}p{0.05\textwidth}p{0.89\textwidth}@{}}
\toprule
\multicolumn{2}{@{}p{0.94\textwidth}@{}}{\textbf{Algorithm \themethodalg: Semantic Segment-Level Memory Construction}} \\
\midrule
\multicolumn{2}{@{}p{0.94\textwidth}@{}}{\textbf{Input:} conversation stream $\mathcal{C}=\{x_1,\ldots,x_T\}$; session identifier $s$; memory store $\mathcal{M}$; boundary threshold $\delta$.} \\
\multicolumn{2}{@{}p{0.94\textwidth}@{}}{\textbf{Output:} updated memory store $\mathcal{M}$.} \\
\midrule
1 & Initialize active segment $S\leftarrow\emptyset$, same-session reference context $\rho_s\leftarrow\emptyset$, and local surprise history $H_s\leftarrow\emptyset$. \\
2 & \textbf{Procedure} \textsc{FinalizeSegment}$(S,\mathrm{reason})$: \\
3 & \quad Encode $S$ with reference context $\rho_s$: $(\mathcal{R}, d)\leftarrow\textsc{Encode}(S,\rho_s)$. \\
4 & \quad Normalize each $r\in\mathcal{R}$ into $(\mathrm{id},\mathrm{type},\mathrm{text},\mathrm{entities},\mathrm{tags},\mathrm{temporal},\mathrm{provenance})$. \\
5 & \quad Insert each normalized record into $\mathcal{M}$. \\
6 & \quad Build entity, topic, entity-topic, temporal, and event-frame evidence nodes from record metadata. \\
7 & \quad Update $\rho_s$ with $d$ and recent same-session record summaries; reset $S\leftarrow\emptyset$. \\
8 & \textbf{For each} incoming exchange $x_t\in\mathcal{C}$ \textbf{do}: \\
9 & \quad Index $x_t$ as raw dialogue evidence and embed $x_t$ with the exchange encoder. \\
10 & \quad \textbf{If} $S=\emptyset$, start $S$ with $x_t$ and finalize it if a hard token or exchange-count cap is reached. \\
11 & \quad Compute semantic surprise, cohesion drop, length pressure, and exchange-count pressure for $x_t$ relative to $S$. \\
12 & \quad Combine the boundary signals into $p_t$ and update $H_s$ with the current semantic surprise. \\
13 & \quad \textbf{If} appending $x_t$ would exceed the hard segment capacity, call \textsc{FinalizeSegment}$(S,\mathrm{capacity\_limit})$ and start a new $S$ with $x_t$. \\
14 & \quad \textbf{Else if} $p_t\geq\delta$, call \textsc{FinalizeSegment}$(S,\mathrm{semantic\_boundary})$ and start a new $S$ with $x_t$. \\
15 & \quad \textbf{Else} append $x_t$ to $S$ and finalize $S$ if a hard token or exchange-count cap is reached. \\
16 & \textbf{End for}. If the session is flushed and $S\neq\emptyset$, call \textsc{FinalizeSegment}$(S,\mathrm{session\_flush})$. \\
17 & \textbf{Return} $\mathcal{M}$. \\
\bottomrule
\end{tabular}
}
\end{table*}

\begin{table*}[t]
\centering
\refstepcounter{methodalg}\label{alg:retrieval-text}
{\small
\setlength{\tabcolsep}{3pt}
\begin{tabular}{@{}p{0.05\textwidth}p{0.89\textwidth}@{}}
\toprule
\multicolumn{2}{@{}p{0.94\textwidth}@{}}{\textbf{Algorithm \themethodalg: Plan-Guided Multi-Route Retrieval}} \\
\midrule
\multicolumn{2}{@{}p{0.94\textwidth}@{}}{\textbf{Input:} user query $q$; recent dialogue context $H_q$; memory store $\mathcal{M}$; requested evidence budget $k$; configured planning depth $k_{\mathrm{plan}}$.} \\
\multicolumn{2}{@{}p{0.94\textwidth}@{}}{\textbf{Output:} evidence context $E_q$ for answer generation.} \\
\midrule
1 & Generate a structured retrieval plan with one LLM call: $\Pi\leftarrow\textsc{Plan}(q,H_q)$. \\
2 & Normalize $\Pi$ and select a retrieval strategy from the planned question type. \\
3 & Set the effective evidence budget $k\leftarrow\max(k,k_{\mathrm{plan}},1)$. \\
4 & \textbf{For each} planned route $R_i\in\{R_1,\ldots,R_m\}$ \textbf{do}: \\
5 & \quad Build route-specific query variants from $q$ and $R_i$; initialize candidate set $D_i\leftarrow\emptyset$. \\
6 & \quad \textbf{If} $R_i$ or $\Pi$ specifies a temporal filter, retrieve matching records through temporal evidence nodes and add them to $D_i$. \\
7 & \quad \textbf{For each} query variant $v$ in $R_i$ \textbf{do}: \\
8 & \quad\quad Retrieve entity, topic, entity-topic, and event-frame evidence nodes by semantic search. \\
9 & \quad\quad Expand matched evidence nodes to linked memory records and add them to $D_i$. \\
10 & \quad\quad Retrieve memory records by semantic search over record embeddings and add direct hits to $D_i$. \\
11 & \quad\quad Retrieve raw dialogue turns by semantic search and add raw-turn hits to $D_i$. \\
12 & \quad Merge duplicate candidates inside $D_i$. \\
13 & \quad Apply question-type-specific score adjustments and local evidence expansion. \\
14 & \quad Optionally rerank $D_i$ with a non-generative cross-encoder reranker. \\
15 & \quad Sort $D_i$ to obtain a route-level ranked list $L_i$. \\
16 & \textbf{End for}. Fuse route-level lists $\{L_1,\ldots,L_m\}$ with reciprocal-rank fusion. \\
17 & Preserve route coverage, apply diversity-aware selection, and enforce temporal/source-type budget controls. \\
18 & Serialize selected memory records and raw dialogue snippets into $E_q$. \\
19 & \textbf{Return} $E_q$. \\
\bottomrule
\end{tabular}
}
\end{table*}

\subsection{Plan-Guided Multi-Route Retrieval}
\label{subsec:plan-guided-multi-route-retrieval}

The final component retrieves evidence for a query using one planning step followed by deterministic recall. Long-term memory questions are often heterogeneous: a temporal question may require date filtering, a multi-hop question may require evidence from multiple entities, and a preference question may require stable user facts rather than recent events. A single embedding query cannot express these distinct evidence requirements. Iterative LLM retrieval can address the issue, but it increases query-time token cost. \method{} separates query understanding from evidence collection: the LLM is used once to specify recall routes, while the subsequent retrieval operations are non-generative.

Given a query $q$ and recent dialogue context $H_q$, the planner outputs a structured plan
\begin{equation}
\Pi(q,H_q)=\bigl(y,\{R_1,\ldots,R_m\}\bigr),
\tag{8}
\end{equation}
where $y$ is the question type and each route $R_j=(g_j,Q_j,C_j,T_j)$ contains a route goal, one or more search queries, structured constraints, and optional temporal constraints. The question type selects retrieval parameters such as channel weights, per-channel candidate limits, and expansion depth.

Each route executes four recall channels in parallel. Direct record recall searches the record vector store using route-specific search queries. Evidence-node recall searches entity, topic, entity-topic, and event-frame nodes, then expands matched nodes to their linked records. Temporal recall applies exact date or range constraints over temporal nodes. Raw-turn recall searches verbatim dialogue embeddings to recover evidence that may not have been consolidated into typed records. Each candidate receives channel-specific scores and retains its provenance.

Candidates from all routes are merged with reciprocal-rank fusion:
\begin{equation}
\mathrm{RRF}(d)=\sum_{j=1}^{m}\frac{1}{\kappa+\mathrm{rank}_j(d)},
\tag{9}
\end{equation}
where $\mathrm{rank}_j(d)$ is the rank of candidate $d$ under route $R_j$ and $\kappa$ is a smoothing constant. Candidates are optionally reranked by a lightweight cross-encoder within each route before route-level lists are fused. The fused candidates are then passed through diversity-aware selection so that evidence from different routes is preserved. The selected records and raw-turn snippets are serialized with their memory types, temporal scopes, and provenance metadata to form the final context for answer generation. When retrieved evidence conflicts, the answer model is instructed to prefer the most recent supported information.

After evidence selection, the answer model generates the final response. Query-time token cost includes both planning and answer generation.

\providecommand{\bluebf}[1]{\ifdefined\textcolor\textcolor{blue}{\textbf{#1}}\else\textbf{#1}\fi}

\section{Experiments}
\label{sec:experiments}

\subsection{Experimental Setup}
\label{subsec:experimental-setup}

\paragraph{Datasets.}
We evaluate \method{} on two long-term conversational memory benchmarks that cover different interaction patterns. \textbf{LoCoMo}~\citep{locomo} focuses on companion-style casual sharing scenarios, comprising 10 multi-session conversations (averaging ${\sim}600$ turns, ${\sim}16$K tokens) with 1,986 question-answer pairs in total. Following the standard long-term memory QA setting, we retain 1,540 questions from the single-hop, multi-hop, temporal-reasoning, and open-domain categories to test whether memory systems can recover fine-grained evidence from cross-session history. \textbf{LongMemEval-S}~\citep{longmemeval} focuses on agent-style task-oriented interactions with significantly longer dialogue contexts (500 conversations, averaging ${\sim}115$K tokens), covering six categories: user facts, assistant facts, preference tracking, multi-session reasoning, knowledge update, and temporal reasoning, posing stricter challenges for memory construction and retrieval under long-term high-load scenarios. Together, these benchmarks cover the core scenario we address: dialogue history grows continuously, and systems must preserve usable evidence under limited construction and query budgets. Detailed benchmark composition is provided in Appendix~\ref{app:benchmark-data}.

\paragraph{Baselines.}
We compare against baselines covering different design directions. \textbf{Full Context} places the entire dialogue history directly into the LLM context window, serving as a direct long-context reference without external memory. \textbf{Naive RAG} chunks the conversation and retrieves via embedding similarity, representing standard retrieval-augmented approaches. Among dedicated long-term memory systems, \textbf{Mem0}~\citep{mem0} invokes the LLM after each turn for fact extraction and update, representing the standard eager consolidation paradigm; \textbf{A-Mem}~\citep{a-mem} extends this with Zettelkasten-style linked notes and dynamic memory evolution, representing an agentic memory design with linked and evolving records; \textbf{MemoryOS}~\citep{memoryos} employs an OS-inspired layered memory architecture (short-term / mid-term / long-term persona modules), representing more complex memory organization directions. Additionally, we compare with \textbf{MemOS}~\citep{memos}, \textbf{Nemori}~\citep{nemori}, \textbf{LightMem}~\citep{LightMem}, and \textbf{TiMem}~\citep{timem} on both benchmarks, and with \textbf{MemU}~\citep{memu} on LoCoMo.

\paragraph{Implementation.}
We evaluate \method{} using \texttt{GPT-4.1-Mini} and \texttt{GPT-4o-Mini} as backbones. Each backbone is used throughout the pipeline for memory encoding, query planning, and final answer generation, with temperature set to 0. Embeddings are computed with \texttt{text-embedding-3-small}, and retrieval-side reranking uses \texttt{bge-reranker-v2-m3}. Additional implementation and prompt details are provided in Appendices~\ref{app:implementation} and~\ref{app:prompts}.

\paragraph{Answer and judge protocol.}
Within each backbone setting, all methods use the same benchmark split and judge protocol. For \method{}, the selected backbone is also used for memory encoding and query planning. Outputs in both settings are evaluated by \texttt{GPT-4o-Mini} using benchmark-specific judge prompts. On LoCoMo, the judge receives the question, gold answer, and generated answer and accepts semantically equivalent phrasings; category accuracy is computed within each retained question type, and overall accuracy over all 1,540 retained questions. On LongMemEval-S, we use the official task-specific yes/no judge with a 10-token output cap. Preference questions use rubric-style desired responses. For knowledge-update questions, the judge accepts the updated answer even if previous information is also mentioned; for temporal-reasoning questions, it applies the official duration tolerance. Judge calls are excluded from all token counts.

\paragraph{Metrics.}
We report overall accuracy and fine-grained category accuracy following each benchmark's official categorization. Overall accuracy is a micro-average over the valid evaluation set: 1,540 retained questions for LoCoMo and 500 questions for LongMemEval-S, with no additional filtering.

\paragraph{Token accounting.}
For the efficiency analysis, we use \texttt{GPT-4.1-Mini} and count construction and query tokens as the sum of generative LLM input and output tokens. For every method, construction tokens cover all generative memory-building calls, whereas query tokens cover final answer generation and any method-specific query-time generative operations. For \method{}, construction tokens comprise the segment-encoding prompt, finalized segment text, reference/disambiguation context, memory-record output, and disambiguation-state output; query tokens comprise the query-planner and final answer-generation input/output, including serialized evidence and recent dialogue context. Embedding computation, deterministic indexing, retrieval and filtering, reciprocal-rank fusion, non-generative reranking, diversity-aware selection, and judge calls are excluded; final answer generation is excluded from construction but included in query cost. Construction cost is averaged per conversation or conversation-question instance, query cost per question, and both are reported in thousands (K). Reporting both quantities separates write-side from query-side cost and reveals whether gains merely shift computation between stages.

\subsection{Main Results}
\label{subsec:main-results}

\begin{table*}[t]
\centering
\small
\setlength{\tabcolsep}{7pt}
\renewcommand{\arraystretch}{1.08}

\caption{Main results on LoCoMo with \texttt{GPT-4.1-Mini} and \texttt{GPT-4o-Mini}. Accuracy (\%) is reported for the official question types and the overall score; higher is better. Best results for each model are bold, and \method{} results are shown in blue.}
\label{tab:locomo_main}

\begin{tabular*}{\linewidth}{@{}c|c|@{\extracolsep{\fill}}ccccc@{}}
\toprule
\textbf{Backbone} & \textbf{Method}
& \multicolumn{1}{c}{\textbf{Single Hop}}
& \multicolumn{1}{c}{\textbf{Multi Hop}}
& \multicolumn{1}{c}{\textbf{Temporal}}
& \multicolumn{1}{c}{\textbf{Open Domain}}
& \multicolumn{1}{c}{\textbf{Overall}} \\
\midrule
\multirow{11}{*}{\rotatebox[origin=c]{90}{\textbf{GPT-4.1-Mini}}}
& Full Context & 90.84 & 82.62 & 79.13 & 57.29 & 84.80 \\
& Naive RAG & 61.24 & 58.87 & 33.96 & 50.00 & 54.42 \\
& MemOS & 85.37 & 79.43 & 75.08 & 64.58 & 80.84 \\
& MemU & 74.91 & 72.34 & 43.61 & 54.17 & 66.62 \\
& MemoryOS & 77.05 & 66.31 & 47.66 & 55.21 & 67.60 \\
& Mem0 & 66.23 & 58.16 & 63.86 & 44.79 & 62.92 \\
& A-Mem & 73.25 & 59.93 & 72.90 & 42.71 & 68.83 \\
& Nemori & 84.90 & 75.10 & 77.60 & 51.00 & 79.40 \\
& LightMem & 81.57 & 62.77 & 64.17 & 54.17 & 72.79 \\
& TiMem & 87.99 & 78.37 & 84.74 & 59.38 & 83.77 \\
& \bluebf{\method{}} & \bluebf{93.34} & \bluebf{87.23} & \bluebf{86.60} & \bluebf{67.71} & \bluebf{89.22} \\
\midrule
\multirow{11}{*}{\rotatebox[origin=c]{90}{\textbf{GPT-4o-Mini}}}
& Full Context & \textbf{90.01} & \textbf{69.86} & 52.96 & 55.21 & 76.43 \\
& Naive RAG & 57.55 & 54.61 & 24.30 & 51.04 & 49.68 \\
& MemOS & 81.45 & 69.15 & 72.27 & \textbf{60.42} & 75.97 \\
& MemU & 72.77 & 62.41 & 33.96 & 46.88 & 61.17 \\
& MemoryOS & 74.55 & 58.87 & 44.24 & 46.88 & 63.64 \\
& Mem0 & 65.52 & 52.84 & 52.02 & 37.50 & 58.64 \\
& A-Mem & 66.83 & 52.13 & 61.99 & 30.21 & 60.84 \\
& Nemori & 82.10 & 65.30 & 71.00 & 44.80 & 74.40 \\
& LightMem & 76.61 & 67.02 & 76.32 & 45.83 & 72.87 \\
& TiMem & 81.43 & 62.20 & 77.63 & 52.08 & 75.30 \\
& \textcolor{blue}{\textbf{\method{}}} & \textcolor{blue}{84.90} & \textcolor{blue}{68.09} & \bluebf{80.06} & \textcolor{blue}{54.17} & \bluebf{78.90} \\
\bottomrule
\end{tabular*}
\end{table*}

\begin{table*}[t]
\centering
\small
\setlength{\tabcolsep}{5pt}
\renewcommand{\arraystretch}{1.08}

\caption{Main results on LongMemEval-S with \texttt{GPT-4.1-Mini} and \texttt{GPT-4o-Mini}. Accuracy (\%) is reported for each official question category and the overall score; higher is better. Best results for each model are bold, and \method{} results are shown in blue.}
\label{tab:longmem_main}

\begin{tabular*}{\linewidth}{@{}c|c|@{\extracolsep{\fill}}ccccccc@{}}
\toprule
\textbf{Backbone} & \textbf{Method}
& \multicolumn{1}{c}{\textbf{SSU}}
& \multicolumn{1}{c}{\textbf{SSA}}
& \multicolumn{1}{c}{\textbf{SSP}}
& \multicolumn{1}{c}{\textbf{MS}}
& \multicolumn{1}{c}{\textbf{KU}}
& \multicolumn{1}{c}{\textbf{TR}}
& \multicolumn{1}{c}{\textbf{Overall}} \\
\midrule
\multirow{10}{*}{\rotatebox[origin=c]{90}{\textbf{GPT-4.1-Mini}}}
& Full Context & 91.43 & \textbf{100.00} & 56.67 & 53.38 & 75.64 & 48.12 & 66.20 \\
& Naive RAG & 85.71 & 82.36 & 83.33 & 63.91 & 75.64 & 45.86 & 67.20 \\
& MemOS & 90.00 & 64.29 & 50.00 & 54.14 & 70.51 & 63.91 & 65.20 \\
& MemoryOS & 94.29 & 89.29 & \textbf{100.00} & 67.67 & 80.77 & 54.89 & 74.40 \\
& Mem0 & 95.71 & 50.00 & 90.00 & 69.92 & 74.36 & 62.41 & 71.20 \\
& A-Mem & 95.71 & \textbf{100.00} & 63.33 & 61.65 & 82.05 & 52.63 & 71.60 \\
& Nemori & 90.00 & 92.90 & 86.70 & 55.60 & 79.50 & 72.20 & 74.60 \\
& LightMem & 90.00 & 23.21 & 76.67 & 51.13 & 88.46 & 84.21 & 69.60 \\
& TiMem & 92.86 & 78.57 & 73.33 & 66.92 & 79.49 & 72.93 & 75.80 \\
& \textcolor{blue}{\textbf{\method{}}} & \bluebf{100.00} & \textcolor{blue}{98.21} & \textcolor{blue}{90.00} & \bluebf{87.97} & \bluebf{97.44} & \bluebf{87.22} & \bluebf{92.20} \\
\midrule
\multirow{10}{*}{\rotatebox[origin=c]{90}{\textbf{GPT-4o-Mini}}}
& Full Context & 44.29 & 80.36 & 53.33 & 35.34 & 55.13 & 34.59 & 45.60 \\
& Naive RAG & 85.71 & 83.93 & 46.67 & 52.63 & 65.38 & 41.35 & 59.40 \\
& MemOS & 95.71 & 67.86 & \textbf{96.67} & 70.67 & 74.26 & \textbf{77.44} & 77.80 \\
& MemoryOS & \textbf{97.14} & 89.29 & 70.00 & 58.65 & 73.08 & 48.87 & 67.80 \\
& Mem0 & 95.71 & 55.36 & 70.00 & 57.89 & 69.23 & 54.14 & 64.40 \\
& A-Mem & 92.54 & \textbf{98.21} & 36.67 & 53.38 & 70.51 & 44.36 & 63.20 \\
& Nemori & 88.60 & 83.90 & 46.70 & 51.10 & 61.50 & 61.70 & 64.20 \\
& LightMem & 87.14 & 32.14 & 68.18 & 71.74 & 83.12 & 67.18 & 69.81 \\
& TiMem & 95.71 & 82.14 & 63.33 & 70.83 & \textbf{86.16} & 68.42 & 76.88 \\
& \textcolor{blue}{\textbf{\method{}}} & \bluebf{97.14} & \textcolor{blue}{96.43} & \textcolor{blue}{70.00} & \bluebf{72.93} & \textcolor{blue}{74.36} & \textcolor{blue}{72.18} & \bluebf{78.80} \\
\bottomrule
\end{tabular*}
\end{table*}

Tables~\ref{tab:locomo_main} and~\ref{tab:longmem_main} show that \method{} achieves the highest overall accuracy on both benchmarks with either backbone. \method{} reaches 89.22\% on LoCoMo and 92.20\% on LongMemEval-S with \texttt{GPT-4.1-Mini}, compared with 84.80\% for Full Context and 75.80\% for TiMem, the strongest baselines on the respective benchmarks. The corresponding results with \texttt{GPT-4o-Mini} are 78.90\% and 78.80\%, improving over the strongest baselines by 2.47 and 1.00 percentage points. Section~\ref{subsec:efficiency} further compares the accuracy and token costs of MemoryOS, Mem0, A-Mem, TiMem, and \method{} using \texttt{GPT-4.1-Mini}. The consistent gains across both benchmarks indicate that \method{} remains effective with either backbone and across different interaction lengths.

\paragraph{LoCoMo.}
On LoCoMo, \method{}'s largest gains over A-Mem with \texttt{GPT-4.1-Mini} appear in multi-hop reasoning (87.23\% vs. 59.93\%, +27.3 pp) and open-domain questions (67.71\% vs. 42.71\%, +25.0 pp). It also reaches 93.34\% on single-hop questions, +20.1 pp above A-Mem (73.25\%). On temporal reasoning, \method{} reaches 86.60\%, +13.7 pp above A-Mem (72.90\%) and +38.9 pp above MemoryOS (47.66\%). \method{} achieves 78.90\% overall with \texttt{GPT-4o-Mini}, 18.06 pp above A-Mem and 2.47 pp above Full Context, together with 80.06\% on temporal reasoning. These gains show that \method{} remains effective across direct recall, evidence composition, temporal reasoning, and open-ended questions; Section~\ref{subsec:ablation} further examines the corresponding design choices. Full Context remains competitive on LoCoMo, but with either backbone, its accuracy decreases on the substantially longer conversations in Table~\ref{tab:longmem_main}.

\paragraph{LongMemEval-S.}
On LongMemEval-S, \method{}'s three largest gains over A-Mem with \texttt{GPT-4.1-Mini} occur in temporal reasoning (87.22\% vs. 52.63\%, +34.59 pp), preference tracking (90.00\% vs. 63.33\%, +26.67 pp), and multi-session reasoning (87.97\% vs. 61.65\%, +26.32 pp). It also reaches 97.44\% on knowledge-update questions, compared with 82.05\% for A-Mem, showing that the pipeline can recover updated evidence from the retained history under the benchmark's answer-level criterion~\citep{longmemeval}. On single-session user and assistant facts, \method{} reaches 100.00\% and 98.21\%, respectively. \method{} achieves the highest overall accuracy of 78.80\% with \texttt{GPT-4o-Mini}, 1.00 pp above MemOS, together with the highest multi-session score (72.93\%) and a tied best result on single-session user facts (97.14\%). These results show consistent gains across temporal, preference-intensive, and cross-session evidence needs with both models. MemoryOS reaches 100.00\% on preference tracking with \texttt{GPT-4.1-Mini}, higher than \method{}'s 90.00\%, suggesting that dedicated user-profile modules may provide advantages for preference-intensive queries.

\subsection{Accuracy-Cost Trade-off}
\label{subsec:efficiency}

\begin{figure}[t]
\centering
\includegraphics[width=\linewidth]{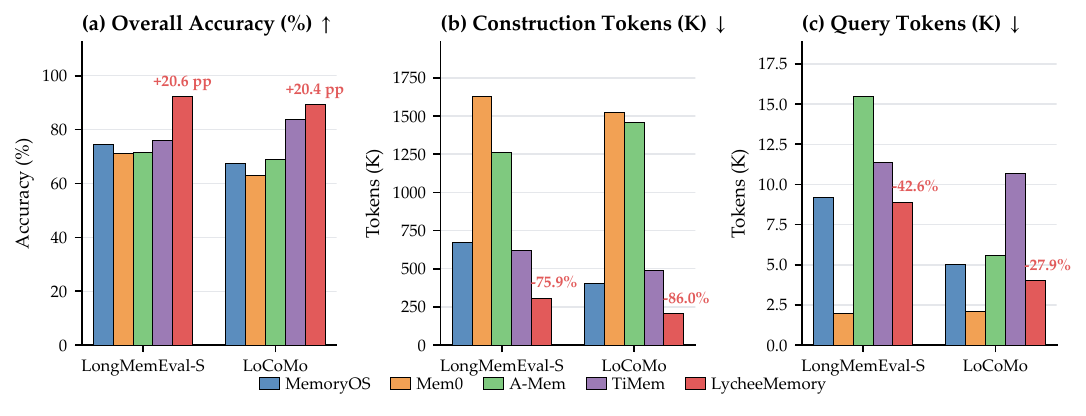}
\caption{Accuracy-cost comparison using \texttt{GPT-4.1-Mini}. The panels report overall accuracy, construction tokens, and query tokens on LongMemEval-S and LoCoMo. \method{} is highlighted in red; annotations show its changes relative to A-Mem.}
\label{fig:accuracy_cost_bars}\end{figure}

Using \texttt{GPT-4.1-Mini}, we compare accuracy, construction tokens, and query tokens in Figure~\ref{fig:accuracy_cost_bars}.

\paragraph{Construction cost.}
On LoCoMo, \method{} uses only 204.1K construction tokens, 86.0\% and 86.6\% lower than A-Mem's 1459.9K and Mem0's 1520.8K respectively, and 58.3\% lower than TiMem's 489.5K. On LongMemEval-S, \method{} uses 304.7K, 75.9\% lower than A-Mem's 1264.3K and 50.9\% lower than TiMem's 620.9K. Construction savings come from segment-level batching: it reduces LLM encoding calls from $T$ (one per turn) to $|\mathcal{S}|$ (one per segment, averaging 5.8 turns per segment on LoCoMo), while each call processes multiple exchanges rather than an individual turn. Semantic boundary detection determines which exchanges are batched together; its contribution relative to fixed-window batching is evaluated separately in Section~\ref{subsec:ablation}.

\paragraph{Query cost.}
\method{} uses fewer query tokens than A-Mem and TiMem on both benchmarks. On LoCoMo, \method{} uses 4.01K query tokens, lower than A-Mem's 5.56K (-27.9\%) and TiMem's 10.71K (-62.6\%). On LongMemEval-S, it uses 8.88K, lower than A-Mem's 15.46K (-42.6\%) and TiMem's 11.36K (-21.8\%). Its retrieval pipeline requires only one LLM planning call; the remaining recall, scoring, fusion, and selection operations are embedding lookups and arithmetic computations that consume no generative tokens.

\paragraph{Overall trade-off.}
Mem0 has the lowest query-token cost but also the lowest accuracy in this comparison, while MemoryOS combines relatively low construction cost with lower accuracy than \method{}.

\subsection{Ablation Study}
\label{subsec:ablation}

Unless otherwise stated, all ablation experiments are conducted on LoCoMo using \texttt{GPT-4.1-Mini}. We assess the contributions of major design choices and component groups to the accuracy-cost trade-off following the system pipeline: construction triggering (Table~\ref{tab:ablation_construction}), memory representation and cross-segment continuity (Table~\ref{tab:ablation_representation}), and query-time retrieval (Table~\ref{tab:ablation_retrieval}). We further vary the boundary threshold used by online semantic segmentation to assess parameter sensitivity (Figure~\ref{fig:threshold_sensitivity}). Detailed variant definitions are provided in Appendix~\ref{app:ablation-definitions}.

\begin{table}[t]
\centering
\caption{Construction-side ablation on LoCoMo with \texttt{GPT-4.1-Mini}. Eager construction removes segment-level batching, whereas fixed-window construction preserves batching but replaces semantic boundary detection with mechanical windows; higher accuracy and lower construction tokens are better.}
\label{tab:ablation_construction}
\begin{tabular}{lrrrrrc}
\toprule
Variant & Overall & Single & Multi & Temp. & Open & Const. $\downarrow$ \\
\midrule
\bluebf{Full \method{}} & \bluebf{89.22} & \bluebf{93.34} & \bluebf{87.23} & \bluebf{86.60} & \bluebf{67.71} & 204.1 \\
eager cons. & 81.88 & 87.16 & 74.47 & 81.31 & 59.38 & 849.9 \\
fixed-window cons. & 82.40 & 90.49 & 74.82 & 75.70 & 56.25 & \textbf{174.7} \\
\bottomrule
\end{tabular}
\end{table}

\paragraph{Construction triggering.}
Replacing segment-level batching with eager construction causes accuracy to drop from 89.22\% to 81.88\% (-7.3 pp) while construction tokens increase from 204.1K to 849.9K (+316\%). Fixed-window consolidation retains batching and uses slightly fewer construction tokens than the full system (174.7K vs. 204.1K), but its accuracy drops to 82.40\% (-6.8 pp), with the largest losses on multi-hop (87.23\% $\to$ 74.82\%) and open-domain questions (67.71\% $\to$ 56.25\%). Together, these comparisons show that batching reduces construction cost, while semantic boundary detection improves accuracy relative to mechanical windows.

\begin{table}[t]
\centering
\caption{Representation-side ablation on LoCoMo with \texttt{GPT-4.1-Mini}. Summary-level records replace typed records, and the context variant removes cross-segment reference feedback; higher accuracy and lower construction tokens are better.}
\label{tab:ablation_representation}
\begin{tabular}{lrrrrrc}
\toprule
Variant & Overall & Single & Multi & Temp. & Open & Const. $\downarrow$ \\
\midrule
\bluebf{Full \method{}} & \bluebf{89.22} & \bluebf{93.34} & \bluebf{87.23} & \bluebf{86.60} & \bluebf{67.71} & 204.1 \\
summary records & 80.78 & 88.35 & 74.47 & 73.83 & 56.25 & \textbf{99.7} \\
w/o cross-seg. context & 81.56 & 87.87 & 75.18 & 76.95 & 60.42 & 189.6 \\
\bottomrule
\end{tabular}
\end{table}

\begin{table}[H]
\centering
\caption{Retrieval-side ablation on LoCoMo with \texttt{GPT-4.1-Mini}. Record-vector only jointly removes structured and raw-turn recall, while w/o fusion/rerank/div jointly removes the selection modules. Higher accuracy and lower query tokens are better.}
\label{tab:ablation_retrieval}
\begin{tabular}{lrrrrrc}
\toprule
Variant & Overall & Single & Multi & Temp. & Open & Query $\downarrow$ \\
\midrule
\bluebf{Full \method{}} & \bluebf{89.22} & \bluebf{93.34} & \bluebf{87.23} & \bluebf{86.60} & \bluebf{67.71} & 4.01 \\
record-vector only & 81.75 & 88.23 & 76.60 & 78.19 & 52.08 & 4.09 \\
w/o query planner & 83.38 & 90.01 & 79.08 & 78.50 & 54.17 & \textbf{2.26} \\
w/o fusion/rerank/div & 66.62 & 71.34 & 60.64 & 66.04 & 44.79 & 3.35 \\
\bottomrule
\end{tabular}
\end{table}

\paragraph{Memory representation.}
Replacing typed, self-contained records with summary-level records reduces construction tokens to 99.7K, while accuracy drops to 80.78\% (-8.4 pp) and temporal reasoning falls from 86.60\% to 73.83\% (-12.8 pp). Removing cross-segment reference context leaves construction tokens nearly unchanged (189.6K vs. 204.1K) but reduces accuracy to 81.56\% (-7.7 pp).

\paragraph{Retrieval pipeline.}
Using record-vector retrieval only, query tokens remain comparable to the full system (4.09K vs. 4.01K) but accuracy drops to 81.75\% (-7.5 pp). Removing the query planner reduces query tokens to 2.26K but accuracy drops to 83.38\% (-5.8 pp). Jointly disabling fusion, reranking, and diversity-aware selection reduces accuracy from 89.22\% to 66.62\% (-22.6 pp).

\begin{figure}[t]
\centering
\includegraphics[width=0.48\linewidth]{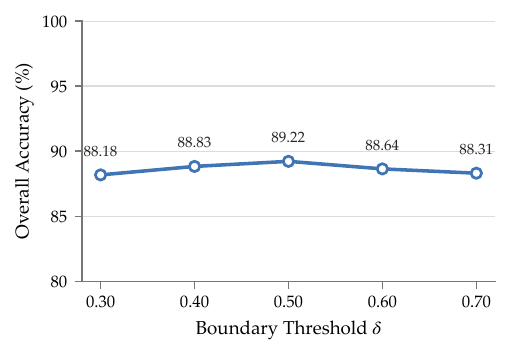}
\caption{Boundary-threshold sensitivity on LoCoMo with \texttt{GPT-4.1-Mini}. We vary the cut threshold $\delta$ applied to $p_t$ while keeping all other settings fixed. The default $\delta=0.50$ achieves the highest accuracy, and overall accuracy varies by 1.04 percentage points across the evaluated range.}
\label{fig:threshold_sensitivity}
\end{figure}

\paragraph{Boundary-threshold sensitivity.}
To assess the stability of online semantic segmentation with respect to its cut criterion, we vary the threshold $\delta$ applied to the boundary probability $p_t$ from 0.30 to 0.70 on the same LoCoMo evaluation set while keeping all other settings fixed. As shown in Figure~\ref{fig:threshold_sensitivity}, overall accuracy ranges from 88.18\% to 89.22\%, a total variation of 1.04 percentage points. The default threshold of 0.50 achieves the highest accuracy; the neighboring settings of 0.40 and 0.60 remain within 0.39 and 0.58 percentage points of the default, respectively. These results show limited sensitivity to the boundary threshold within the evaluated range.

\paragraph{Summary.}
The ablations show accuracy losses or cost increases when construction, representation, or retrieval components are removed or replaced. Across the tested boundary thresholds, accuracy varies by 1.04 percentage points.

\section{Conclusion}
\label{sec:conclusion}

We introduced \method{}, a cost-efficient long-term memory framework for LLM agents that replaces turn-level consolidation with semantic segment-level consolidation. Segment-level batching reduces construction frequency, while semantic boundary detection determines coherent segment composition before each segment is encoded into context-independent typed records. Together with lightweight cross-segment disambiguation, structured evidence indexes, and query-planned multi-route recall, this design preserves fine-grained conversational evidence while avoiding unnecessary construction and query-time overhead. Experiments on LoCoMo and LongMemEval-S demonstrate that this design improves long-term memory QA and yields a stronger accuracy--cost trade-off than representative memory baselines. Future work will integrate user-profile modeling with segment-level memory to address the preference-recall gap observed on LongMemEval-S.


\bibliographystyle{lychee}
\bibliography{custom}

\newpage

\appendix
\appendix

\section{Experimental Protocol Details}
\label{app:experimental-protocol}

This appendix supplements the evaluation setup in Section~\ref{subsec:experimental-setup}. The main text defines the answer-generation and judge protocol, metric computation, and token-accounting boundaries. Here we report benchmark composition, the exact evaluation-component configuration, formal metric definitions, and averaging equations, followed by implementation and prompt details needed for reproduction. Unless otherwise stated, all generation and judge calls use deterministic decoding with temperature set to 0.

\subsection{Benchmark Data and Evaluation Tasks}
\label{app:benchmark-data}

LoCoMo and LongMemEval-S use different evaluation units and question taxonomies. LoCoMo evaluates multiple question-answer pairs per multi-session conversation, whereas LongMemEval-S evaluates one question per long conversation-question instance.

\subsubsection{LoCoMo}
\label{app:locomo}

LoCoMo~\cite{locomo} is a long-context conversational memory benchmark built around companion-style multi-session dialogues. Each dialogue contains two named speakers and requires a memory system to recover fine-grained personal facts, resolve speaker-specific references, and combine evidence distributed across turns or sessions.

\paragraph{Data statistics.}
LoCoMo is small in the number of conversations but dense in annotated memory questions. It contains 10 multi-session conversations, with each conversation averaging approximately 600 turns and 16K tokens. The benchmark provides 1,986 question-answer pairs in total; the main evaluation retains 1,540 QA pairs from the standard long-term memory categories after excluding the adversarial/counterfactual subset.

\paragraph{Task categories.}
The LoCoMo main evaluation uses 1,540 valid question-answer pairs from the first four categories. We exclude category 5 adversarial/counterfactual questions because these examples are not part of the standard semantic-answering setting used for the main long-term memory QA comparison. The same 1,540 questions are used for the main results, cost analysis, and LoCoMo ablations, so method variants are compared on an identical evaluation set. Table~\ref{tab:app_locomo_categories} gives the category-level composition.

\begin{table*}[t]
\centering
\caption{LoCoMo question categories used in the main evaluation. Counts sum to the 1,540 valid QA pairs used for all LoCoMo accuracy and cost analyses.}
\label{tab:app_locomo_categories}
{\small
\setlength{\tabcolsep}{3pt}
\begin{tabular}{@{}p{0.17\textwidth}rp{0.32\textwidth}p{0.36\textwidth}@{}}
\toprule
Question type & \# Questions & Evidence pattern & Capability stressed \\
\midrule
Single-hop & 841 & A localized fact, event, preference, or detail is sufficient to answer the question. & Fine-grained fact preservation and speaker/entity disambiguation. \\
Multi-hop & 282 & Evidence must be combined across multiple turns, events, or sessions. & Cross-session evidence coverage and entity continuity. \\
Temporal reasoning & 321 & The answer depends on dates, ordering, duration, recurrence, or relative time expressions. & Temporal normalization, date-aware retrieval, and reasoning over event order. \\
Open-domain & 96 & The question is phrased more openly and may have weak lexical overlap with the relevant dialogue evidence. & Robust evidence selection under non-template queries. \\
\midrule
Total & 1,540 & All retained LoCoMo QA pairs. & Overall micro-average. \\
\bottomrule
\end{tabular}
}
\end{table*}

\subsubsection{LongMemEval-S}
\label{app:longmemeval}

LongMemEval-S~\cite{longmemeval} focuses on agent-style task-oriented interactions with substantially longer dialogue histories. In contrast to LoCoMo, each evaluation item corresponds to one long conversation-question instance. This setting stresses whether a system can recover user facts, assistant-side facts, preferences, cross-session evidence, changed information, and temporal relations under much higher context load.

\paragraph{Data statistics.}
LongMemEval-S is organized at the instance level rather than as multiple QA pairs over the same conversation. The official cleaned split contains 500 conversation-question instances, with one question attached to each long dialogue history. Its average context length is approximately 115K tokens per instance, making it substantially longer than LoCoMo and better suited for testing memory construction under high context load.

\paragraph{Task categories.}
The LongMemEval-S evaluation uses all 500 questions in the official cleaned split. We report category accuracy using the official question type field and compute the overall score as a micro-average over all 500 questions. For compact table layout, Table~\ref{tab:app_longmem_categories} abbreviates the official category names as SSU (\texttt{single-session-user}), SSA (\texttt{single-session-assistant}), SSP (\texttt{single-session-preference}), MS (\texttt{multi-session}), KU (\texttt{knowledge-update}), and TR (\texttt{temporal-reasoning}).

\begin{table*}[t]
\centering
\caption{LongMemEval-S question categories used in the main evaluation. Counts sum to the 500 questions in the official cleaned split; abbreviations are defined in the text.}
\label{tab:app_longmem_categories}
{\small
\setlength{\tabcolsep}{3pt}
\begin{tabular}{@{}p{0.10\textwidth}rp{0.28\textwidth}p{0.48\textwidth}@{}}
\toprule
Question type & \# Questions & Memory ability & Evidence requirement \\
\midrule
SSU & 70 & User-fact recall & Recover a fact provided by the user within a single session. \\
SSA & 56 & Assistant-side recall & Recover information, commitments, or responses previously given by the assistant. \\
SSP & 30 & Preference tracking & Use user-specific preference information to satisfy the reference rubric. \\
MS & 133 & Cross-session reasoning & Integrate evidence distributed across multiple sessions. \\
KU & 78 & Changed-state recall & Answer with the updated state from information that changes across the dialogue history. \\
TR & 133 & Temporal reasoning & Interpret order, duration, relative time, or time-bounded evidence. \\
\midrule
Total & 500 & Official cleaned split & Overall micro-average. \\
\bottomrule
\end{tabular}
}
\end{table*}

\subsection{Evaluation Component Configuration}
\label{app:judge-protocol}

The answer-generation and judge protocol is defined in Section~\ref{subsec:experimental-setup}. Table~\ref{tab:app_eval_components} records the exact models used by \method{} for memory encoding, query planning, answer generation, embedding, reranking, and benchmark-specific judging. Complete generation and judge prompts are provided in Appendix~\ref{app:prompts}.

\begin{table*}[t]
\centering
\caption{Evaluation components used by \method{}. The same model is used for memory encoding, query planning, and answer generation; the cost and ablation experiments use \texttt{GPT-4.1-Mini}. Temperature is set to 0 for all generative components.}
\label{tab:app_eval_components}
{\small
\setlength{\tabcolsep}{3pt}
\begin{tabular}{@{}p{0.22\textwidth}p{0.27\textwidth}p{0.45\textwidth}@{}}
\toprule
Component & Model & Role \\
\midrule
Memory encoder & \texttt{GPT-4.1-Mini} / \texttt{GPT-4o-Mini} & Convert finalized segments into typed memory records. \\
Query planner & \texttt{GPT-4.1-Mini} / \texttt{GPT-4o-Mini} & Produce typed recall routes for evidence retrieval. \\
Answer generator & \texttt{GPT-4.1-Mini} / \texttt{GPT-4o-Mini} & Generate the final answer from the retrieved evidence. \\
Embedding model & \texttt{text-embedding-3-small} & Embed exchanges, records, and queries. \\
Reranker & \texttt{bge-reranker-v2-m3} & Rerank retrieved evidence candidates without generative decoding. \\
LoCoMo judge & \texttt{GPT-4o-Mini} & Judge semantic correctness for LoCoMo answers. \\
LongMemEval-S judge & \texttt{GPT-4o-Mini} & Apply the official task-specific yes/no judge. \\
\bottomrule
\end{tabular}
}
\end{table*}

\subsection{Formal Metric Definitions}
\label{app:metrics}

Section~\ref{subsec:experimental-setup} defines the valid evaluation sets, accuracy aggregation, and token-accounting boundaries used in the main results. Table~\ref{tab:app_metrics} provides a compact formal summary of the four reported metrics. The exact averaging equations and component-level token inclusions are given in Appendix~\ref{app:token-accounting}.

\begin{table*}[t]
\centering
\caption{Metric definitions and accounting rules. Token counts include generative LLM input and output tokens and are reported in thousands (K).}
\label{tab:app_metrics}
{\small
\setlength{\tabcolsep}{3pt}
\begin{tabular}{@{}p{0.20\textwidth}p{0.45\textwidth}p{0.30\textwidth}@{}}
\toprule
Metric & Definition & Notes \\
\midrule
Category accuracy & Number of correct answers in a category divided by the number of valid questions in that category. & Reported using each benchmark's question taxonomy. \\
Overall accuracy & Number of correct answers divided by the number of valid questions in the benchmark. & Micro-average over questions, not a macro-average over categories. \\
Construction tokens & Generative LLM input and output tokens consumed during memory construction. & Averaged per conversation or conversation-question instance; reported in K. \\
Query tokens & Generative LLM input and output tokens consumed during answer generation and any other query-time generative operations. & Averaged per question; reported in K. \\
\bottomrule
\end{tabular}
}
\end{table*}

\section{Implementation Details}
\label{app:implementation}

This appendix provides the default implementation parameters following the pipeline order in Section~\ref{sec:method}. The four stages are online semantic segmentation, segment-level memory encoding, structured evidence organization, and plan-guided multi-route retrieval.

\subsection{Online Semantic Segmentation}
\label{app:segmentation-details}

Online semantic segmentation decides when a buffered dialogue segment should be encoded by the memory encoder. Unlike turn-level eager consolidation, segment-level batching allows multiple exchanges to share one encoding call and therefore controls construction frequency. Within this batching scheme, the semantic boundary policy determines which exchanges form each segment: \method{} finalizes the active segment when it becomes saturated, a topic transition is detected, or a hard capacity limit is reached. Boundary scoring uses embeddings and deterministic computation only.

Each exchange is embedded with \texttt{text-embedding-3-small}. The system maintains the active segment centroid, the most recent exchange embedding, the local semantic-surprise history, the current token length, and the exchange count. Table~\ref{tab:app_segmentation_params} lists the default segmentation parameters. Let $z_s=(s-0.20)/0.14$ denote the normalized absolute-surprise score. The implementation converts the combined boundary score into a cut probability:

\begin{equation*}
p_t=\sigma(-1.10+\mathrm{score}),
\qquad
\mathrm{cut}\ \text{if}\ p_t\geq 0.50.
\end{equation*}

\begin{table*}[t]
\centering
\caption{Default segmentation parameters. The segmentation step uses no generative LLM calls.}
\label{tab:app_segmentation_params}
{\small
\setlength{\tabcolsep}{3pt}
\begin{tabular}{@{}p{0.26\textwidth}p{0.24\textwidth}p{0.42\textwidth}@{}}
\toprule
Parameter / signal & Value & Function \\
\midrule
Exchange embedding model & \texttt{text-embedding-3-small} & Embed each incoming exchange. \\
Cut probability threshold & 0.50 & Finalize the active segment when the cut probability reaches this threshold. \\
Surprise history window & 64 values & Support local normalization of semantic surprise. \\
Minimum history for robust signal & 5 values & Disable robust surprise normalization when the history is too short. \\
Robust surprise normalization & median/MAD z-score clipped to [-2.0, 4.0] & Reduce outlier and early-stage noise. \\
Absolute surprise signal & $\min\{2.5,\max\{-1.0,z_s\}\}$ & Provide an absolute novelty component. \\
Minimum chunk tokens & 300 & Avoid excessive fragmentation. \\
Target chunk tokens & 600 & Preferred consolidation scale. \\
Maximum chunk tokens & 900 & Hard token cap. \\
Maximum exchanges & 10 & Hard exchange-count cap. \\
\bottomrule
\end{tabular}
}
\end{table*}

The length signal is a piecewise pressure function over the current segment length. It is set to -1.30 below 70\% of the minimum chunk length, grows from -0.80 to 0 before the minimum length, grows from 0.45 to 1.90 between the minimum and target lengths, grows from 1.90 to 2.80 between the target and maximum lengths, and becomes 3.00 at or beyond the maximum length. The turn-count signal is -0.85 for one exchange, -0.15 for two exchanges, 0.15 for three exchanges, and $\min\{1.0,0.30+0.15(n-4)\}$ for four or more exchanges, where $n$ is the current exchange count. If appending a new exchange would exceed the maximum token cap, the current segment is finalized before the new exchange is added.

\subsection{Segment-Level Memory Encoding}
\label{app:encoding-details}

Segment-level memory encoding converts each finalized segment into typed records that can be retrieved and interpreted without the original dialogue context. The encoder receives the current segment text and a compact same-session reference context, then returns memory records and an updated disambiguation state. Tables~\ref{tab:app_encoding_settings} and~\ref{tab:app_record_schema} summarize the encoding settings and record schema.

\begin{table*}[t]
\centering
\caption{Segment-level memory encoding settings.}
\label{tab:app_encoding_settings}
{\small
\setlength{\tabcolsep}{3pt}
\begin{tabular}{@{}p{0.30\textwidth}p{0.58\textwidth}@{}}
\toprule
Item & Setting \\
\midrule
Encoder model & \texttt{GPT-4.1-Mini} or \texttt{GPT-4o-Mini}, matching the answer model \\
Temperature & 0 \\
Max output tokens & Provider default \\
Output format & Raw JSON object without markdown fences \\
\texttt{disambiguation\_context} budget & 1,200 characters \\
Same-session reference context budget & 2,400 characters \\
Recent semantic records retained & 12 \\
\bottomrule
\end{tabular}
}
\end{table*}

\begin{table*}[t]
\centering
\caption{Memory record fields produced by the segment encoder.}
\label{tab:app_record_schema}
{\small
\setlength{\tabcolsep}{3pt}
\begin{tabular}{@{}p{0.24\textwidth}p{0.64\textwidth}@{}}
\toprule
Field & Description \\
\midrule
\texttt{record\_id} & Internal identifier used to link storage and retrieval entries. \\
\texttt{memory\_type} & One of \texttt{fact}, \texttt{preference}, \texttt{event}, \texttt{constraint}, \texttt{procedure}, \texttt{failure\_pattern}, or \texttt{tool\_affordance}. \\
\texttt{semantic\_text} & Self-contained natural-language memory statement. \\
\texttt{normalized\_text} & Concatenation of memory type and semantic text. \\
\texttt{entities} & Canonical entity names. \\
\texttt{tags} & Topic, action, or type tags. \\
\texttt{temporal} & Normalized event or validity times, represented by fields such as \texttt{t\_ref}, \texttt{t\_valid\_from}, and \texttt{t\_valid\_to}. \\
\texttt{evidence\_turn\_range} & Source turn indexes. \\
\texttt{source\_session} & Source session identifier. \\
\texttt{source\_role} & \texttt{user}, \texttt{assistant}, \texttt{both}, or empty. \\
\texttt{confidence} & Current implementation uses 1.0. \\
\bottomrule
\end{tabular}
}
\end{table*}

\subsection{Structured Evidence Organization}
\label{app:organization-details}

Structured evidence organization converts metadata from the segment encoder into searchable evidence nodes. In addition to record embeddings, \method{} organizes records into entity, topic, entity-topic, temporal, and event-frame indexes. This phase does not call a generative LLM; it uses embedding, SQLite/FTS indexing, and deterministic bookkeeping. Table~\ref{tab:app_organization_settings} summarizes the organization settings.

\begin{table*}[t]
\centering
\caption{Structured evidence organization settings.}
\label{tab:app_organization_settings}
{\small
\setlength{\tabcolsep}{3pt}
\begin{tabular}{@{}p{0.24\textwidth}p{0.28\textwidth}p{0.38\textwidth}@{}}
\toprule
Component & Implementation & Notes \\
\midrule
Vector backend & LanceDB & Stores record embeddings for direct semantic retrieval. \\
Structured store & SQLite + FTS5 & Stores evidence nodes and full-text searchable fields. \\
Entity/tag normalization & Case folding, separator normalization, punctuation cleanup, repeated-whitespace merge & Produces stable keys. \\
Temporal nodes & Day-level and month-level keys & Derived from \texttt{t\_ref}, \texttt{t\_valid\_from}, and \texttt{t\_valid\_to}. \\
Event-frame index & Finalized semantic segment & Each node groups the records produced from one finalized semantic segment and retains its source session and turn range as provenance. \\
Evidence-node merge & \texttt{node\_type:key} & Merges repeated entity, topic, and temporal nodes. \\
\bottomrule
\end{tabular}
}
\end{table*}

\method{} retains textually distinct statements as separate records, including records that may describe successive or conflicting states. The organizer does not perform approximate semantic merging, infer supersession links, close the validity interval of an earlier record, or automatically expire it when a later statement is inserted. This preserves the available historical evidence while leaving conflict resolution to retrieval and answer generation.

\subsection{Plan-Guided Multi-Route Retrieval}
\label{app:retrieval-details}

Plan-guided multi-route retrieval uses one LLM planning call for query understanding and then executes non-generative evidence collection. The planner's internal question types are \texttt{single}, \texttt{aggregate}, \texttt{temporal}, \texttt{comparison}, \texttt{personalized\_advice}, \texttt{prior\_assistant\_response}, and \texttt{other}. These types are not benchmark categories; they select retrieval strategies, candidate budgets, and route constraints. Table~\ref{tab:app_retrieval_params} lists the default retrieval parameters. Let $k_{\mathrm{req}}$ denote the requested evidence budget, $k_{\mathrm{plan}}$ the planning depth, and $k$ the effective final budget. We use $\lambda_{\mathrm{node}}$, $\lambda_{\mathrm{record}}$, and $\lambda_{\mathrm{turn}}$ for the question-type-specific recall multipliers.

\begin{table*}[t]
\centering
\caption{Default retrieval parameters.}
\label{tab:app_retrieval_params}
{\small
\setlength{\tabcolsep}{3pt}
\begin{tabular}{@{}p{0.34\textwidth}p{0.54\textwidth}@{}}
\toprule
Retrieval parameter & Value \\
\midrule
Planner model & \texttt{GPT-4.1-Mini} or \texttt{GPT-4o-Mini}, matching the answer model; temperature 0 \\
Final top-$k$ & $k=\max\{k_{\mathrm{req}},k_{\mathrm{plan}},1\}$ \\
Planning depth $k_{\mathrm{plan}}$ & 15 \\
Evidence-node recall budget & $\lambda_{\mathrm{node}}\max\{3k,30\}$ \\
Direct record recall budget & $\lambda_{\mathrm{record}}\max\{3k,30\}$ \\
Raw-turn recall budget & $\lambda_{\mathrm{turn}}\max\{k,20\}$ \\
Temporal filter recall budget & $\max\{10k,100\}$ \\
RRF smoothing constant & 60.0 \\
Reranker & \texttt{bge-reranker-v2-m3} \\
Route-level rerank candidate limit & $\max\{c_{\mathrm{rerank}},4k\}$ \\
Configured reranker candidate limit $c_{\mathrm{rerank}}$ & 100 \\
\bottomrule
\end{tabular}
}
\end{table*}

Each route can activate direct record recall, evidence-node recall, temporal recall, and raw-turn recall. Route-level candidates are merged with reciprocal-rank fusion using a smoothing constant of 60.0. After fusion, the system first preserves route coverage using a per-route quota of $\max\{1,\min\{4,\lfloor k/m\rfloor\}\}$, where $m$ is the number of planned routes. The final candidate score is:

\begin{equation*}
s_{\mathrm{final}}
=0.82\,s_{\mathrm{best}}
+0.18\,s_{\mathrm{route}}.
\end{equation*}

Diversity-aware selection uses an MMR-style objective:

\begin{equation*}
s_{\mathrm{select}}
=0.75\,s_{\mathrm{rel}}
-0.25\,s_{\mathrm{sig}}.
\end{equation*}

The candidate signature contains the source session, entities, matched queries, evidence nodes, turns, and text tokens. The final evidence budget is count-based top-$k$ rather than a fixed token budget, so query prompt length varies with the selected evidence text. This is why Section~\ref{subsec:efficiency} reports empirical average query tokens.

\section{Token Accounting Details}
\label{app:token-accounting}

This appendix formalizes the token-accounting boundaries defined in Section~\ref{subsec:experimental-setup}. The distinction between construction tokens and query tokens is central to the paper's conclusion: \method{} aims to reduce write-side construction cost without shifting the cost to query-time context expansion.

\subsection{Construction Tokens}
\label{app:construction-tokens}

Construction tokens count all recorded generative LLM input and output tokens consumed during memory building, averaged by conversation or conversation-question instance and reported in thousands (K):

\begin{equation*}
\mathrm{ConstructionTokens}
=
\frac{
\sum_{c\in\mathcal{C}_{\mathrm{build}}}
\left(t^{\mathrm{in}}_c+t^{\mathrm{out}}_c\right)
}{
N_{\mathrm{conv}}
}.
\end{equation*}

For \method{}, construction tokens include all recorded memory-encoding input and output tokens, covering the segment memory encoding prompt, finalized segment text, reference/disambiguation context, memory-record output, and disambiguation-state output. Construction tokens exclude embedding computation, deterministic structured indexing, SQLite/FTS/vector operations, BM25 or vector search, non-generative reranking, evaluation judge calls, and final answer generation calls.

\subsection{Query Tokens}
\label{app:query-tokens}

Query tokens count all recorded generative LLM input and output tokens consumed while answering evaluation questions, averaged by question and reported in thousands (K):

\begin{equation*}
\mathrm{QueryTokens}
=
\frac{
\sum_{q\in\mathcal{Q}}
\left(t^{\mathrm{in}}_q+t^{\mathrm{out}}_q\right)
}{
|\mathcal{Q}|
}.
\end{equation*}

For \method{}, query tokens include the query-planner input/output, final answer-generation input/output, serialized evidence context, and recent dialogue context included in the answer prompt. Query tokens exclude embedding retrieval, structured filtering, reciprocal-rank fusion, non-generative cross-encoder reranking, diversity-aware selection, and evaluation judge calls. Because the final evidence context is count-based rather than token-budget based, query tokens reflect the average prompt length after selected evidence has been serialized.

\section{Ablation Variant Definitions}
\label{app:ablation-definitions}

This appendix defines the implementation of each ablation variant in Section~\ref{subsec:ablation}. The main text discusses the results, while the appendix specifies which modules are removed, replaced, or preserved in each comparison.

\subsection{Construction-Side Variants}
\label{app:construction-variants}

\paragraph{per-turn construction (w/o segment-level batching).}
This variant removes segment-level batching and falls back to turn-level eager construction. On LoCoMo, it triggers memory construction at the speaker-turn ingestion granularity; on LongMemEval-S, it uses the original message/exchange granularity. The record schema, structured indexing, and query-time retrieval pipeline remain unchanged. This variant evaluates the effect of replacing segment-level batching with eager construction.

\paragraph{fixed-window consolidation.}
This variant replaces semantic boundary detection with fixed-window batching. It preserves batching, but segment boundaries no longer depend on semantic surprise or cohesion drop. The default configuration uses 600 target chunk tokens and at most 10 exchanges. This variant evaluates semantic boundary detection relative to mechanical windows at a comparable batching scale.

\subsection{Representation-Side Variants}
\label{app:representation-variants}

\paragraph{summary-level records.}
This variant replaces typed, self-contained records with a summary-level representation. It no longer explicitly retains the typed atomic structure formed by memory type, entity, topic, temporal scope, and provenance fields, resulting in summary-level retrieval without these structured fields. The rest of the retrieval pipeline remains unchanged, enabling a comparison between typed records and summary-level memory.

\paragraph{w/o cross-segment reference context.}
This variant encodes each segment independently and does not pass same-session disambiguation context or recent resolved records to subsequent segments. The record schema and retrieval pipeline remain unchanged. This variant evaluates the cross-segment context configuration.

\subsection{Retrieval-Side Variants}
\label{app:retrieval-variants}

\paragraph{record-vector retrieval only.}
This variant disables entity, topic, temporal, event-frame, and raw-turn multi-route recall, and uses only memory-record vector search to return evidence candidates. The final evidence budget is kept consistent with the requested top-$k$ of the full system. This variant evaluates the combined structured and raw-turn recall channels relative to memory-record vector search.

\paragraph{w/o query planner.}
This variant removes the LLM query planner and constructs a fixed single-route query directly from the original question. It performs no LLM query rewriting, question-type classification, or route decomposition. Query tokens therefore decrease, but retrieval routes cannot adapt to temporal, comparison, personalized-advice, or prior-assistant-response intents.

\paragraph{w/o fusion/reranking/diversity selection.}
This variant preserves recall candidates but jointly removes reciprocal-rank fusion, cross-encoder reranking, and diversity-aware selection. Candidates are sorted by their original retrieval or field scores and truncated to top-$k$. This variant evaluates the combined evidence-selection stack.

\section{Prompt Templates and Evaluation Prompts}
\label{app:prompts}

This appendix reports the prompt templates used by \method{} and the evaluation prompts used for judging. The prompt templates below are those used in our implementation for memory encoding, query planning, and answer generation. The LongMemEval-S judge prompts follow the official task-specific yes/no evaluator, and the LoCoMo judge prompt follows the JSON label protocol used in the LoCoMo evaluation implementation. Prompt blocks below are line-wrapped for typesetting.

\subsection{\method{} Segment Encoding Prompt}
\label{app:segment-encoding-prompt}

Segment encoding uses a system message and a user message. The user message contains \texttt{<SESSION\_DATE>}, \texttt{<REFERENCE\_CONTEXT>}, and \texttt{<CURRENT\_TURNS>}. The reference context is used only for resolving references and aliases; it is not treated as a source from which new facts may be extracted.

\begin{PromptBox}{Segment Encoding Prompt --- System Message}
You are a memory encoder. Given a conversation chunk, extract atomic memory records that are self-contained and retrievable without the original dialogue.

\PromptHeading{\#\# Input}

\begin{itemize}
\tightlist
\item
  \texttt{\textless{}SESSION\_DATE\textgreater{}}: date anchor for converting relative dates. May be absent.
\item
  \texttt{\textless{}REFERENCE\_CONTEXT\textgreater{}}: notes from earlier chunks in the same session. Use only to resolve references; never extract facts from it.
\item
  \texttt{\textless{}CURRENT\_TURNS\textgreater{}}: the turns to extract from. \texttt{evidence\_turns} are 0-based indexes into this block.
\end{itemize}

\PromptHeading{\#\# Rules}

\begin{enumerate}
\def\labelenumi{\arabic{enumi}.}
\tightlist
\item
  One fact per record. Write each \texttt{semantic\_text} as a standalone sentence - replace every pronoun, alias, and implicit subject with the explicit referent. Preserve exact names, quantities, dates, places, reasons, and outcomes.
\item
  Convert relative dates ("yesterday", "next Monday") to absolute dates (YYYY-MM-DD) in both \texttt{semantic\_text} and \texttt{temporal} when \texttt{\textless{}SESSION\_DATE\textgreater{}} is provided.
\item
  Conversation chunks may contain explicit speaker prefixes inside the content, such as "Gina: ..." or "Jon: ...", even when the outer turn role is \texttt{user}. Treat those prefixed names as the true speakers.
\item
  When a named speaker says "I", "me", or "my", resolve it to that speaker\textquotesingle s name. Do not rewrite named speakers as "the user". Use "the user" only when no speaker name is available.
\item
  Store durable facts, preferences, plans, constraints, and events stated by any real conversation speaker. Skip greetings, filler, acknowledgements, generic encouragement, and generic world knowledge.
\item
  Store assistant content only when it is a substantive remembered part of the conversation; skip ingestion acknowledgements such as "OK" or "Acknowledged".
\item
  Return \texttt{disambiguation\_context} as a short note listing resolved speaker names, aliases, relationships, and dates needed for later reference resolution. Empty string if none.
\item
  If no durable information exists, return \{"records": {[}{]}, "disambiguation\_context": ""\}.
\end{enumerate}

\PromptHeading{\#\# Output - raw JSON, no markdown}

\begin{Verbatim}[breaklines=true,breakanywhere=true,fontsize=\footnotesize]
{
  "records": [
    {
      "memory_type":
        "fact | preference | event | constraint | procedure |
         failure_pattern | tool_affordance",
      "semantic_text": "complete standalone sentence",
      "entities": ["searchable named entity"],
      "temporal": {"t_ref": "", "t_valid_from": "", "t_valid_to": ""},
      "tags": ["short keyword"],
      "evidence_turns": [0],
      "source_role": "user | assistant | both"
    }
  ],
  "disambiguation_context": ""
}
\end{Verbatim}

\textbf{Memory types:} fact (stable attribute/relationship/state); preference

(like/dislike/habit/recurring choice); event (dated one-off occurrence); constraint (limit/policy/restriction); procedure (reusable steps/workflow); failure\_pattern (mistake/pitfall/lesson); tool\_affordance (tool capability or limitation).
\end{PromptBox}

\begin{PromptBox}{Segment Encoding Prompt --- User Message}
\begin{Verbatim}[breaklines=true,breakanywhere=true,fontsize=\footnotesize]
<SESSION_DATE>{session_date}</SESSION_DATE>
<REFERENCE_CONTEXT>
{reference_context}
</REFERENCE_CONTEXT>

<CURRENT_TURNS>
{current_turns}
</CURRENT_TURNS>
\end{Verbatim}
\end{PromptBox}

\subsection{\method{} Query Planning Prompt}
\label{app:query-planning-prompt}

The query planning prompt converts the current question into an executable retrieval plan. The planner describes only the evidence need visible in the user question and recent context; it does not decide whether the answer exists and does not invent answer candidates. For ordinary named-speaker factual questions, the default behavior is a single route; multiple routes are used only when the question explicitly requires separate evidence.

\begin{PromptBox}{Query Planning Prompt --- System Message}
Write a JSON plan for answering a user question from past conversation memory.

\textbf{Inputs:}

\begin{itemize}
\tightlist
\item
  \texttt{\textless{}USER\_QUERY\textgreater{}}: current request.
\item
  \texttt{\textless{}RECENT\_CONTEXT\textgreater{}}: recent turns, if any.
\end{itemize}

\textbf{Your job:}

\begin{itemize}
\tightlist
\item
  Look only at the user-visible wording in \texttt{\textless{}USER\_QUERY\textgreater{}} and \texttt{\textless{}RECENT\_CONTEXT\textgreater{}}.
\item
  Describe only what the user is asking for.
\item
  Do not decide whether the answer is available or unavailable.
\item
  Do not speculate beyond the user-visible question.
\end{itemize}

\textbf{Question classification:}

\begin{itemize}
\tightlist
\item
  \texttt{question\_type} describes the question form visible from the request: \texttt{single}, \texttt{aggregate}, \texttt{temporal}, \texttt{comparison}, \texttt{personalized\_advice}, \texttt{prior\_assistant\_response}, or \texttt{other}.
\item
  Some questions ask about one named speaker in a two-speaker conversation. Use \texttt{single} for ordinary who / what / where / why / how factual questions about a speaker, event, preference, plan, object, or image detail.
\item
  Use \texttt{temporal} only when the final answer needs a date, approximate date, order, duration, recurrence, or a relative-time conversion.
\item
  Use \texttt{comparison} only when the question directly compares two or more non-time attributes.
\item
  Use \texttt{aggregate} only when the question asks for a count, list, set, total, or all matching items.
\item
  Use \texttt{personalized\_advice} only for advice or recommendation questions.
\item
  Use \texttt{prior\_assistant\_response} only when the question asks what the assistant previously said.
\end{itemize}

\textbf{Query generation:}

\begin{itemize}
\tightlist
\item
  Produce 3-8 short semantic queries. Prefer high-precision phrases over broad paraphrases.
\item
  Preserve exact speaker names, people, places, and object names from the question. Do not rewrite named speakers as "the user".
\item
  Include the named speaker with the key predicate in most queries.
\item
  Add 1-2 natural paraphrases for likely conversation wording, but do not invent answer entities.
\item
  Use the same language as the user query when natural.
\end{itemize}

\textbf{Evidence target and constraints:}

\begin{itemize}
\tightlist
\item
  \texttt{evidence\_target} is the single fact/event/detail/set needed to answer the question.
\item
  \texttt{evidence\_constraints} should stay short: named speaker, key predicate, requested attribute, and any explicit time/status condition.
\item
  Do not add generic hidden requirements that are not visible in the user question.
\end{itemize}

\textbf{Evidence routes:}

\begin{itemize}
\tightlist
\item
  Use one \texttt{evidence\_routes} item by default. The route should retrieve the most relevant dialogue span or memory for the named speaker and predicate.
\item
  Use multiple routes only when the question explicitly needs separate evidence: two named speakers, two events being compared, two endpoints of a duration, or a count/list over several members.
\item
  Do not create separate routes for generic helper facts such as "age basis", "background", "current state", or "candidate events" unless the question explicitly requires them.
\item
  Each route should have a concise \texttt{evidence\_goal}, 3-5 route-specific \texttt{queries}, optional \texttt{constraints}, and optional \texttt{temporal\_filter}.
\end{itemize}

\textbf{Time:}

\begin{itemize}
\tightlist
\item
  Set \texttt{temporal\_filter} only when the request or recent context contains a concrete date or an unambiguous range that can be converted to YYYY-MM-DD.
\item
  For vague time wording such as "recently", "earlier", "later", "next month", or "last week", usually leave \texttt{temporal\_filter} empty and rely on retrieved conversation timestamps.
\item
  For "on DATE", set both \texttt{since} and \texttt{until} to DATE. For "before DATE", set only \texttt{until}. For "after DATE" or "since DATE", set only \texttt{since}.
\end{itemize}

\textbf{Self-check before returning:}

\begin{itemize}
\tightlist
\item
  The plan should be compact. If a simple named-speaker question has more than one route, merge it.
\item
  If the queries could match the wrong speaker, add the correct speaker name to them.
\item
  If the plan contains invented answer candidates, remove them.
\end{itemize}

\textbf{Return raw JSON only:}

\begin{Verbatim}[breaklines=true,breakanywhere=true,fontsize=\footnotesize]
{
  "reason": "brief planning reason: what the user asks, target/constraints,
             query strategy",
  "question_type":
    "single|aggregate|temporal|comparison|personalized_advice|
     prior_assistant_response|other",
  "semantic_queries": ["query"],
  "temporal_filter": {},
  "evidence_target": "",
  "evidence_constraints": [],
  "constraints": [
    {"kind": "time|entity|status|relation|attribute|other", "value": ""}
  ],
  "evidence_routes": [
    {
      "route_id": "r1",
      "evidence_goal": "specific user-visible evidence this route should find",
      "queries": ["route-specific query"],
      "constraints": [
        {"kind": "time|entity|status|relation|attribute|other", "value": ""}
      ],
      "temporal_filter": {}
    }
  ]
}
\end{Verbatim}
\end{PromptBox}

\subsection{\method{} Answer Generation Prompt}
\label{app:answer-generation-prompt}

The answer generation prompt receives two evidence blocks, episodic/semantic memories and raw memories. The prompt instructs the model to avoid double-counting overlapping evidence, answer only about the named person when a question names a person, and prioritize the most recent supported information when memories conflict.

\begin{PromptBox}{Answer Generation Prompt --- System Message}
You are an intelligent memory assistant tasked with answering questions using conversation memories.

\PromptHeading{\# CONTEXT}

You have access to memories from two speakers in a conversation. These memories are timestamped and may be relevant to the question.

There are two memory blocks:

\begin{enumerate}
\def\labelenumi{\arabic{enumi}.}
\tightlist
\item
  Episodic/Semantic Memories: refined summaries and concise, fact-like pieces extracted from conversations.
\item
  Raw Memories: unprocessed conversation snippets between the two speakers.
\end{enumerate}

\textbf{Context Rules:}

\begin{enumerate}
\def\labelenumi{\arabic{enumi}.}
\tightlist
\item
  Episodic/Semantic memories and Raw memories may overlap in content. Avoid double-counting redundant evidence.
\item
  Carefully analyze both memory blocks and identify information that is actually useful for answering the question.
\item
  Memories are sorted by relevance.
\item
  The conversation may involve two named speakers. When the question names a person, answer about that person only and do not substitute facts about the other speaker.
\item
  In Raw Memories, speaker prefixes inside the text, such as "Gina:" or "Jon:", identify the true speaker. Use those names to resolve "I", "me", and "my".
\item
  A \texttt{Date:} line on an Episodic/Semantic memory is the resolved event or valid date for that memory.
\item
  A \texttt{Mentioned:} line on a Raw Memory is the conversation date when that raw snippet was said.
\end{enumerate}

\PromptHeading{\# INSTRUCTIONS}

\begin{enumerate}
\def\labelenumi{\arabic{enumi}.}
\tightlist
\item
  Carefully read all provided memories.
\item
  Pay close attention to timestamps when time is relevant.
\item
  If the question asks about a specific event or fact, look for direct, explicit evidence in the memories.
\item
  If the question asks for advice, recommendations, or what kind of response the user would prefer, first identify the named person\textquotesingle s preferences, habits, constraints, or past actions, then base the suggestion primarily on these person-specific signals.
\item
  If memories contain contradictory information, prioritize the most recent memory.
\item
  For time references, convert them into concrete dates based on the memory timestamp.
\item
  In raw memories, \texttt{Mentioned:} marks the conversation time. Do not confuse the time of conversation with the time when an event actually happened.
\item
  Do not say "no information found" if there are related memories that can reasonably guide a personalized answer. Only abstain when there is truly no relevant evidence.
\item
  Use exact words or a short phrase from the memories when possible.
\item
  The final answer should usually be less than 5-6 words. Do not include explanation unless needed to avoid ambiguity.
\end{enumerate}

\PromptHeading{\# APPROACH (Think step by step internally)}

\begin{enumerate}
\def\labelenumi{\arabic{enumi}.}
\tightlist
\item
  Identify whether the question is factual or asks for advice/preference.
\item
  Retrieve all memories that are clearly related to the question.
\item
  Check timestamps and content to locate the most reliable information.
\item
  For factual queries, pinpoint explicit mentions that answer the question.
\item
  For advice/preference queries, anchor the answer in person-specific facts.
\item
  If temporal reasoning or calculation is needed, do it internally and convert the result into a concrete date or time span.
\item
  Formulate a precise, concise answer supported by the memories.
\item
  Output only the answer phrase.
\end{enumerate}

\{time\_basis\}

\textbf{Episodic/Semantic Memories:}

\{episodic\_semantic\_memories\}

\textbf{Raw Memories:}

\{raw\_memories\}

\textbf{Question:} \{question\}

\textbf{Answer:}
\end{PromptBox}

\subsection{LongMemEval-S Official Judge Prompts}
\label{app:longmem-judge-prompts}

The official LongMemEval-S evaluator selects task-specific judge prompts for standard, temporal-reasoning, knowledge-update, preference, and abstention questions. We use \texttt{GPT-4o-Mini} as the judge with temperature 0 and a 10-token output cap; a response containing ``yes'' is parsed as correct.

\begin{PromptBox}{LongMemEval-S Judge Prompt --- Standard}
I will give you a question, a correct answer, and a response from a model. Please answer yes if the response contains the correct answer. Otherwise, answer no. If the response is equivalent to the correct answer or contains all the intermediate steps to get the correct answer, you should also answer yes. If the response only contains a subset of the information required by the answer, answer no.

\textbf{Question:} \{question\}

\textbf{Correct Answer:} \{answer\}

\textbf{Model Response:} \{response\}

Is the model response correct? Answer yes or no only.
\end{PromptBox}

\begin{PromptBox}{LongMemEval-S Judge Prompt --- Temporal Reasoning}
I will give you a question, a correct answer, and a response from a model. Please answer yes if the response contains the correct answer. Otherwise, answer no. If the response is equivalent to the correct answer or contains all the intermediate steps to get the correct answer, you should also answer yes. If the response only contains a subset of the information required by the answer, answer no. In addition, do not penalize off-by-one errors for the number of days. If the question asks for the number of days/weeks/months, etc., and the model makes off-by-one errors, the model\textquotesingle s response is still correct.

\textbf{Question:} \{question\}

\textbf{Correct Answer:} \{answer\}

\textbf{Model Response:} \{response\}

Is the model response correct? Answer yes or no only.
\end{PromptBox}

\begin{PromptBox}{LongMemEval-S Judge Prompt --- Knowledge Update}
I will give you a question, a correct answer, and a response from a model. Please answer yes if the response contains the correct answer. Otherwise, answer no. If the response contains some previous information along with an updated answer, the response should be considered as correct as long as the updated answer is the required answer.

\textbf{Question:} \{question\}

\textbf{Correct Answer:} \{answer\}

\textbf{Model Response:} \{response\}

Is the model response correct? Answer yes or no only.
\end{PromptBox}

\begin{PromptBox}{LongMemEval-S Judge Prompt --- Preference}
I will give you a question, a rubric for desired personalized response, and a response from a model. Please answer yes if the response satisfies the desired response. Otherwise, answer no. The model does not need to reflect all the points in the rubric. The response is correct as long as it recalls and utilizes the user\textquotesingle s personal information correctly.

\textbf{Question:} \{question\}

\textbf{Rubric:} \{answer\}

\textbf{Model Response:} \{response\}

Is the model response correct? Answer yes or no only.
\end{PromptBox}

\begin{PromptBox}{LongMemEval-S Judge Prompt --- Abstention}
I will give you an unanswerable question, an explanation, and a response from a model. Please answer yes if the model correctly identifies the question as unanswerable. The model could say that the information is incomplete, or some other information is given but the asked information is not.

\textbf{Question:} \{question\}

\textbf{Explanation:} \{answer\}

\textbf{Model Response:} \{response\}

Does the model correctly identify the question as unanswerable? Answer yes or no only.
\end{PromptBox}

\subsection{LoCoMo Accuracy Metric and Judge Prompt}
\label{app:locomo-judge-prompt}

LoCoMo uses accuracy as the primary metric. For each question, GPT-4o-Mini judges whether the generated answer is semantically consistent with the gold answer. Samples labeled \texttt{CORRECT} count as correct; category and overall accuracy are then computed over the retained valid questions.

\begin{PromptBox}{LoCoMo Judge Prompt --- System Message}
You are an expert grader that determines if answers to questions match a gold standard answer.
\end{PromptBox}

\begin{PromptBox}{LoCoMo Judge Prompt --- User Message}
Your task is to label an answer to a question as \textquotesingle CORRECT\textquotesingle{} or \textquotesingle WRONG\textquotesingle. You will be given the following data:

\begin{enumerate}
\def\labelenumi{\arabic{enumi}.}
\tightlist
\item
  a question (posed by one user to another user),
\item
  a \textquotesingle gold\textquotesingle{} (ground truth) answer,
\item
  a generated answer which you will score as CORRECT/WRONG.
\end{enumerate}

The point of the question is to ask about something one user should know about the other user based on their prior conversations. The gold answer will usually be a concise and short answer that includes the referenced topic, for example:

\textbf{Question:} Do you remember what I got the last time I went to Hawaii?

\textbf{Gold answer:} A shell necklace

The generated answer might be much longer, but you should be generous with your grading - as long as it touches on the same topic as the gold answer, it should be counted as CORRECT.

For time related questions, the gold answer will be a specific date, month, year, etc. The generated answer might be much longer or use relative time references, but you should be generous with your grading - as long as it refers to the same date or time period as the gold answer, it should be counted as CORRECT. Even if the format differs, consider it CORRECT if it is the same date.

Now it\textquotesingle s time for the real question:

\textbf{Question:} \{question\}

\textbf{Gold answer:} \{ground\_truth\}

\textbf{Generated answer:} \{answer\}

First, provide a short (one sentence) explanation of your reasoning, then finish with CORRECT or WRONG.

\textbf{Output strictly in JSON format:}

\begin{Verbatim}[breaklines=true,breakanywhere=true,fontsize=\footnotesize]
{
  "label": "CORRECT" | "WRONG",
  "reasoning": "<short explanation>"
}
\end{Verbatim}
\end{PromptBox}

\section{Limitations and Artifact Use}
\label{app:limitations-artifact}

Our evaluation covers text-only long-term conversational memory and measures construction and query costs in generative LLM tokens. End-to-end latency, storage growth, and performance during continuous deployment remain to be evaluated.

The main results show a remaining weakness on preference-intensive questions. On LongMemEval-S, \method{} reaches 90.00\% with \texttt{GPT-4.1-Mini}, below MemoryOS at 100.00\%, and 70.00\% with \texttt{GPT-4o-Mini}, below MemOS at 96.67\%. This suggests that dedicated persona or user-profile modeling may provide advantages for preference-heavy queries. A future system could combine segment-level evidence construction with stronger profile modeling.

The reported results use hosted LLM and embedding APIs and the \texttt{bge-reranker-v2-m3} reranker specified in Appendix~\ref{app:judge-protocol}. Accuracy and token consumption depend on these model choices.

The LoCoMo copy used in our experiments is licensed under Creative Commons Attribution-NonCommercial 4.0 International (CC BY-NC 4.0), which restricts use to non-commercial purposes and requires attribution. The LongMemEval reference copy is MIT licensed, with redistribution subject to retaining copyright and license notices. The benchmark materials retain their upstream licenses.

\end{document}